%% file: main.tex
\documentclass[screen]{acmart}
\usepackage{amsmath,amsfonts}
\usepackage{algorithmic}
\usepackage{graphicx}
\usepackage{textcomp}
\usepackage{xcolor}
\usepackage{url}
\usepackage{bbm}
\usepackage{booktabs}
\usepackage{mathtools}
\usepackage{wrapfig}
\usepackage[normalem]{ulem}
\usepackage{mathrsfs}
\usepackage{multirow}
\usepackage{pifont}
\usepackage{subfigure}
\usepackage{CJKutf8}
\newcommand{\revision}[1]{\textcolor{black}{#1}}
\AtBeginDocument{%
  \providecommand\BibTeX{{%
    \normalfont B\kern-0.5em{\scshape i\kern-0.25em b}\kern-0.8em\TeX}}}

\setcopyright{acmcopyright}
\copyrightyear{2018}
\acmYear{2018}
\acmDOI{XXXXXXX.XXXXXXX}

\acmConference[Conference acronym 'XX]{Make sure to enter the correct
  conference title from your rights confirmation emai}{June 03--05,
  2018}{Woodstock, NY}
\acmPrice{15.00}
\acmISBN{978-1-4503-XXXX-X/18/06}

\def\model{TIPIN}

\def\fullmodel{Transformer-based Interdisciplinary Path Inference Network}

\def\chillmodel{\textbf{T}ransformer-based \textbf{I}nterdisciplinary \textbf{P}ath \textbf{I}nference \textbf{N}etwork}

\begin{document}

\title[Transformer-based Interdisciplinary Path Inference Network]{Interdisciplinary Fairness in Imbalanced Research Proposal Topic Inference: A Hierarchical Transformer-based Method with Selective Interpolation}

\author{Meng Xiao}
\email{shaow@cnic.cn}
\orcid{0000-0001-5294-5776}
\affiliation{%
  \institution{1.Computer Network Information Center, Chinese Academy of Sciences, Beijing; 2.University of Chinese Academy of Sciences, Beijing}
  \country{China}
}
\author{Min Wu}
\orcid{0000-0003-0977-3600}
\affiliation{%
  \institution{Institute for Infocomm Research, Agency for Science, Technology and Research}
  \country{Singapore}}
\email{wumin@i2r.a-star.edu.sg}

\author{Ziyue Qiao}
\email{ziyuejoe@gmail.com}
\orcid{0000-0002-9485-4861}
\affiliation{%
  \institution{School of Computing and Information Technology, Great Bay University}
  \city{Dongguan}
  \state{Guangdong}
  \country{China}
}

\author{Yanjie Fu}
\affiliation{%
  \institution{Arizona State University, School of Computing and AI, United States}
  \country{USA}}
\email{yanjiefu@asu.edu}
\orcid{0000-0002-1767-8024}

\author{Zhiyuan Ning}
\email{ningzhiyuan@cnic.cn}
\orcid{0000-0003-4852-0163}
\author{Yi Du}
\email{duyi@cnic.cn}
\orcid{0000-0003-4852-0163}
\author{Yuanchun Zhou}
\orcid{0000-0003-2144-1131}
\email{zyc@cnic.cn}
\affiliation{%
   \institution{Computer Network Information Center, Chinese Academy of Sciences; 2.University of Chinese Academy of Sciences, Beijing}
  \city{Beijing}
  \country{China}
}
\thanks{Yi Du and Ziyue Qiao are the corresponding authors.}
\renewcommand{\shortauthors}{Meng et al.}

\ccsdesc[300]{Computing methodologies~Natural language processing}

\input{abstract}


\received{20 February 2007}
\received[revised]{12 March 2009}
\received[accepted]{5 June 2009}

\maketitle


\section{Introduction}
\input{Introduction}

\section{Preliminar}
\input{problem}

\section{Methodology}
\input{method}

\section{Experiment Settings}
\input{experiment_setting}

\section{Experiment Results}
\input{experiment}

\section{Conclusion}
\input{conclusion}

\section{Related Work}
\input{related}

\begin{acks}
This work is partially supported by the Excellent Young Scientists Fund of NSFC (No.T2322027), the Postdoctoral Fellowship Program of CPSF (No.GZC20232736), the China Postdoctoral Science Foundation Funded Project (No.2023M743565), the Special Research Assistant Funded Project of the Chinese Academy of Sciences,
Youth Innovation Promotion Association CAS.
\end{acks}

\bibliographystyle{ACM-Reference-Format}
\bibliography{ref}

\end{document}

%% file: abstract.tex
\begin{abstract}
The objective of topic inference in research proposals aims to obtain the most suitable disciplinary division from the discipline system defined by a funding agency. 
The agency will subsequently find appropriate peer review experts from their database based on this division. Automated topic inference can reduce human errors caused by manual topic filling, bridge the knowledge gap between funding agencies and project applicants, and improve system efficiency. 
Existing methods focus on modeling this as a hierarchical multi-label classification problem, using generative models to iteratively infer the most appropriate topic information. 
However, these methods overlook the gap in scale between interdisciplinary research proposals and non-interdisciplinary ones, leading to an unjust phenomenon where the automated inference system categorizes interdisciplinary proposals as non-interdisciplinary, causing unfairness during the expert assignment. 
How can we address this data imbalance issue under a complex discipline system and hence resolve this unfairness? 
In this paper, we implement a topic label inference system based on a Transformer encoder-decoder architecture. 
Furthermore, we utilize interpolation techniques to create a series of pseudo-interdisciplinary proposals from non-interdisciplinary ones during training based on non-parametric indicators such as cross-topic probabilities and topic occurrence probabilities. 
This approach aims to reduce the bias of the system during model training. 
Finally, we conduct extensive experiments on a real-world dataset to verify the effectiveness of the proposed method. 
The experimental results demonstrate that our training strategy can significantly mitigate the unfairness generated in the topic inference task.
To improve the reproducibility of our research, we have released accompanying code by Dropbox.\footnote{\tiny \url{https://www.dropbox.com/sh/fyex0p1pzyvr9lt/AABLrJy6-dbTTM5cw5PLDMp8a?dl=0}}.
\end{abstract}

%% file: Introduction.tex
The evaluation and approval of research grants is a multifaceted process that involves a diverse range of stakeholders~\cite{qiao2022rpt,qiao2023semi}, including researchers, grant administrators, and expert reviewers. This procedure requires drafting research proposals, organizing peer reviews based on topic relevance and potential conflicts of interest, and evaluating societal benefits and innovation potential.
One of the challenges inherent in this peer-review system is the assignment of proposals to suitable domain-specific reviewers, a step that is critical to ensuring the efficiency and fairness of the review process. Upholding fairness throughout this process is imperative for the robust and healthy advancement of the scientific research community.

The real-world problem of research proposal topic inference faces three primary challenges:
(1) The text of research proposals presents various characteristics, manifesting in different lengths, semantic structures, and roles within the proposal. For instance, an abstract provides an overall description of a proposal's research content. At the same time, the body text delivers a more detailed account, and the title serves as the most concise elaboration of the work.
(2) During fund topic inference, various topics appear in disciplines and are organized into a complex tree structure. This implies that topic inference must consider how to utilize this hierarchical structure and contemplate the granularity of inference during the prediction process.
(3) Interdisciplinary and non-interdisciplinary research is vital in today's grant application process. However, due to the fewer interdisciplinary studies, there is an imbalance between interdisciplinary and non-interdisciplinary research in the actual data set (interdisciplinary-non-interdisciplinary imbalance issue). Training on such an imbalanced dataset could lead to bias in AI models. This may result in interdisciplinary research being assigned to non-interdisciplinary experts unfamiliar with the intersectional area for review, thereby causing serious unfairness issues.

\begin{figure}
\centering
\includegraphics[width=0.5\textwidth]{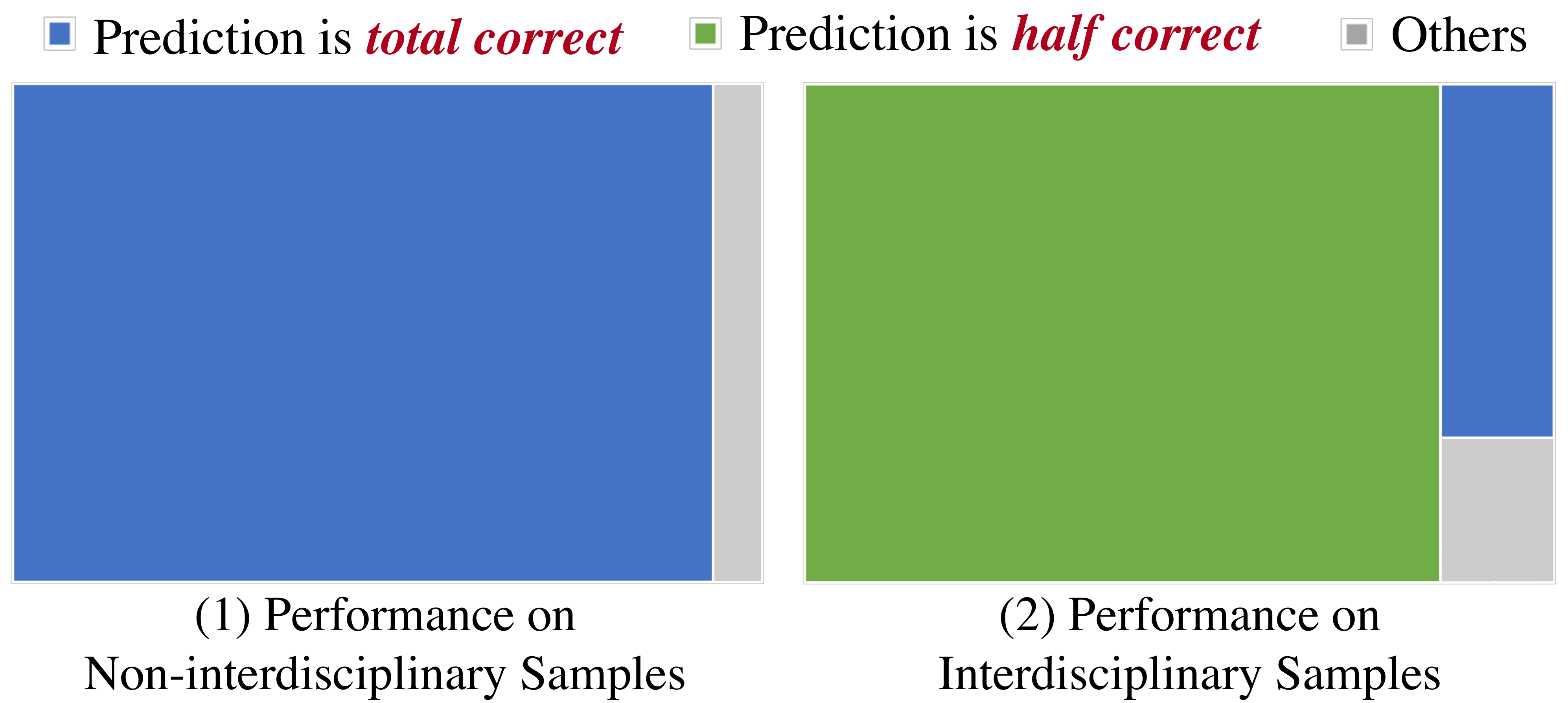}
\caption{An illustration of the consequence of the interdisciplinary-non-interdisciplinary imbalance issue. The HIRPCN performance on the interdisciplinary test set and the non-interdisciplinary test set. (1) the model will perform nicely on non-interdisciplinary samples, and most test samples are correct, i.e., present as the blue shade. (2) most cases are partially correct on interdisciplinary samples due to the interdisciplinary-non-interdisciplinary imbalance issue, i.e., present as the green shade.}
\label{figure.issue}
\vspace{-0.5cm}
\end{figure}

Previous studies~\cite{cai2023resolving} have partially explored these issues. For instance, HMT~\cite{xiao2021expert} constructed a hierarchical topic inference framework, which demonstrated strong performance in single-topic scenarios. And HIRPCN~\cite{xiao2023hierarchical} investigated the application of interdisciplinary knowledge in topic inference tasks, developing a complex method for interdisciplinary topic inference. 
These studies attempted to address the modeling of applications and hierarchical topic inference tasks but did not consider the serious imbalance caused by the interdisciplinary-non-interdisciplinary imbalance issue.
Figure~\ref{figure.issue} demonstrates the detailed categorization effects of HIRPCN on real-world datasets consisting of interdisciplinary and non-interdisciplinary data. It's apparent that the model achieved good results on non-interdisciplinary data sets. However, on the validation set of the interdisciplinary data set, most interdisciplinary proposals were inferred as non-interdisciplinary research. This type of inference result could lead to serious fairness issues during the downstream expert categorization process.

\textbf{Our perspectives: selective interpolation for generating high-quality pseudo-training data.} We define the task of inferring the topic of a proposal as a hierarchical multi-label inference problem, which reduces the number of labels the model has to deal with each time during the inference process, significantly lowering the complexity of model inference. In addition, to address the issue of modeling various types of textual data encompassed by the proposal, we propose a hierarchical Transformer~\cite{ye2023needed} architecture that models different component texts separately. 
During the topic inference process, we consider the process of manual topic inference by experts and design a method to aggregate textual semantic information based on hierarchical inference adaptively. To mitigate potential fairness issues, we design a Selective Interpolation method. This method constructs pseudo-cross features by combining samples with strong interdisciplinarity and challenging prediction characteristics to balance the dataset, thereby reducing the model's bias in inferring the topic of interdisciplinary data.

\textbf{Summary of Contributions.} To address the aforementioned issues, this paper proposes a novel framework, namely \fullmodel, for improving the research proposal modeling, topic inference, and fairness issue in automatic topic inference for research proposal. Our contributions can be summarized as follows: 
(1) We study an interesting AI-assisted proposal classification problem. The proposal data exhibit three fundamental complexities: hierarchical discipline structure, heterogeneous textual semantics, and interdisciplinary-non-interdisciplinary imbalance.
(2) We generalize this proposal classification problem into a hierarchical multi-label path classification task. We propose a novel framework \model\ that enables the integration of transformer-based textual semantics modeling and level-wise prediction architecture, thus advancing the proposal modeling and topic inference.
(3) To alleviate the fairness issue from interdisciplinary-non-interdisciplinary imbalance, we develop a novel selective interpolation method that can select high-quality samples to create pseudo-training data, thus making the model more objective for topic inference.
(4)  We conduct extensive experiments to evaluate the performance of various selective interpolation strategies. In particular, the Word-level MixUp strategy can balance the attention between each keyword from different domains and achieve the best performance compared to other MixUp strategies.

%% file: problem.tex
\begin{definition}[\textbf{Research Proposal}]
A research proposal, denoted as $D$, is a set of documents submitted by applicants for grant considerations. These documents can include elements like \textit{Title}, \textit{Abstract}, \textit{Keywords}, \textit{Research Fields}, among others. The documents within a proposal are represented as $D = \{d_t\}_{t=1}^{|T|}$, where $t$ specifies the type of each document from a set of all possible document types, $T$. Here, $|T|$ is the total count of document types and $d_i$ indicates the document of type $t_i$. Each document $d_i$ is a series of word tokens represented by $d_i = [w^{1}_i, w^{2}_i, ..., w_{i}^{|d_i|}]$, with $w^{k}_i$ being the $k$-th word token in $d_i$.
\end{definition}

\begin{definition}[\textbf{Hierarchical Discipline Structure}]\label{d1}
A hierarchical discipline structure, symbolized as $\gamma$, is either a Directed Acyclic Graph (DAG) or a tree. It consists of discipline entities connected by a directed \textit{Belong-to} relation, indicating a discipline's sub-disciplines.
The set of discipline nodes is represented as $C=\{C_0\cup C_1\cup ...\cup C_H\}$, arranged across $H$ hierarchical levels. Here, $H$ denotes the depth of the hierarchy. Each level $C_k$ is given by $C_k=\{c^i_k\}^{\mid C_k\mid }_{i=1}$, describing the disciplines in the $i$-th level. The root level is denoted by $C_0=\{root\}$, serving as the origin of $\gamma$.
The connection between disciplines is characterized by a partial order, $\prec$, representing the \textit{Belong-to} relationship. This relationship is:
\begin{itemize}
    \item \textbf{Asymmetric}: For all $c^x_i \in C_i$ and $c^y_j \in C_j$, if $c^x_i \prec c^y_j$ then $c^y_j \not\prec c^x_i$.
    \item \textbf{Anti-reflexive}: For any $c^x_i \in C_i$, $c^x_i \not\prec c^x_i$.
    \item \textbf{Transitive}: For all $c^x_i \in C_i$, $c^y_j \in C_j$, and $c^z_k \in C_k$, if $c^x_i \prec c^y_j$ and $c^y_j \prec c^z_k$, then $c^x_i \prec c^z_k$.
\end{itemize}
Conclusively, the Hierarchical Discipline System $\gamma$ is defined as a partial order set $\gamma=(C,\prec)$.
\end{definition}

\begin{definition}[\textbf{Topic Inference Problem}]
The Topic Inference Problem is modeled as a Hierarchical Multi-label Classification task. Within this model, the disciplinary codes of proposals are represented using a sequence of discipline-level-specific label sets, denoted by $\textbf{L} = [L_0, L_1, L_2,..., L_{H_A}]$. Here, $L_0 = \{l_{root}\}$, and $L_i=\{l_i^j\}_{j=1}^{\mid L_i\mid }$ is a set of labels corresponding to the label paths at the $i$-th level, such that $\forall l_i^j \in L_i, l_i^j \in C_i$. The maximum length of these label paths is given by $H_A$.
In this setup, given a proposal's document set $D$, the prediction process is broken down in a top-down approach, moving from the initial level through the hierarchical discipline system $\gamma$. If the labels of the first $k-1$ ancestors in $L$ are denoted by $\textbf{L}_{<k} = [L_{0}, L_1,..., L_{k-1}]$ (with $L_{<1}=\{L_0\}$), then the prediction at level $k$ is essentially a multi-label classification task at that level, formulated as:
$$\Omega(D,\textbf{L}_{<k},\gamma;\Theta) \to \textbf{L}_{k}$$ 
Here, $\Theta$ represents the parameters of model $\Omega$.

Ultimately, the probability of assigning a sequence of label sets to a proposal during prediction can be expressed as:
\begin{equation}
\label{propability}
    P(L\vert D,\gamma;\Theta)=\prod_{k=1}^{H_A} P(L_k\vert D,\textbf{L}_{<k},\gamma; \Theta)
\end{equation}
In this equation, $L_k \subset \textbf{L}$ denotes the target label set at level-$k$. $P(L\vert D,\gamma;\Theta)$ signifies the cumulative probability of a proposal being associated with the label set sequence $\textbf{L}$. Meanwhile, $P(L_k\vert D,\gamma,\textbf{L}_{<k};\Theta)$ is the probability of a research proposal being assigned a label set at level-$k$, given its preceding ancestor $\textbf{L}_{<k}$.

The training objective, provided all the ground truth labels, is to maximize the value of Equation~\ref{propability}.
\end{definition}

%% file: method.tex
\begin{figure*}[!h]
\centering
\includegraphics[width=\linewidth]{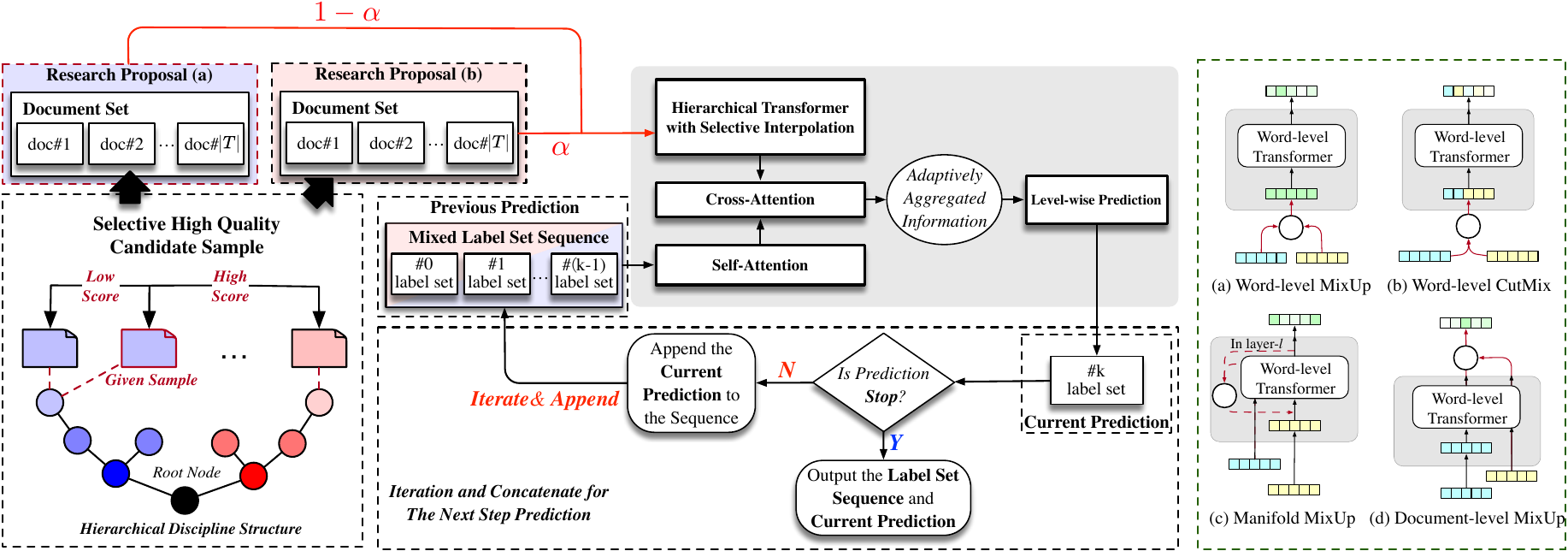}
\caption{An overview of \model.   1) On the left side, the Selective Interpolation will select two high-quality candidate samples to MixUp. 2) The grey shade is the architecture illustration of \model, which will generate the current step's prediction. 3) \model\ will follow the iterative progress shown at the bottom of the figure to continue prediction or stop at the current level. 4) The right side is the demonstration of interpolation strategy variants.}
\label{model_overview}
\end{figure*}

In this section, we focus on presenting the architecture of \chillmodel\ (\model). 
Figure~\ref{model_overview} illustrates the whole pipeline of the proposed method. 
We begin by discussing how to model heterogeneous research proposals, where primary components are the hierarchical Transformer that models different parts of each document separately, thereby acquiring context information at various levels. 
We then discuss how to identify the most suitable objects for interpolation fusion during the training process. 
In the final two subsections,
we explain how topic inference is performed on the Hierarchical Discipline Structure, mainly how to aggregate textual semantic information based on the inference process and how to make hierarchical predictions, as well as how to optimize the model based on the primary pipeline. These subsections constitute the main pipeline of the model.

\subsection{Modeling Heterogeneous Research Proposal}

\noindent\textbf{Why modeling each document separately matters.}
A research proposal represents a complex blend of several documents, each holding unique characteristics. 
Employing conventional neural language processing techniques to merge this textual data could result in significant information loss. 
To effectively represent the research proposal, extracting the information from each document separately and then adaptively integrating them becomes essential. 
We develop a 2-level hierarchical transformer architecture to advance this feature. 
The word-level transformer is designed to model each document independently, while the document-level transformer aspires to aggregate them dependently.

\smallskip
\noindent\textbf{Leveraging hierarchical transformer.}
Now we present the two-level hierarchical transformer framework:

\noindent \textit{\uline{Word-level Transformer:}} aims to extract the semantic information within the input text sequence:
\begin{equation}
\begin{aligned}
    \mathbf{W}^{(l)}_{i} &= \text{Transformer}(\mathbf{W}^{(l-1)}_i),
\end{aligned}
\end{equation}
where the $\text{Transformer}(\cdot)$ is a multi-layer Transformer. The $\mathbf{W}^{(l)}_i \in \mathbb{R}^{n \times h}$ is the intermediate state of the type $i$ document representation in layer-$(l)$ of the Word-level Transformer. The input $\mathbf{W}_{i}^{(0)}$ in the first layer can be initialized with look-up input token sequence $d_i$ via a pre-trained Word2Vec~\cite{mikolov2013distributed} model.

\noindent \textit{\uline{Document-level Transformer:}} aims to utilize the type of document (e.g., the title, the abstract, etc.) to model the importance of the input and generate the type-specific representation of the research proposal:
\begin{equation}
    \begin{aligned}
        \label{trans2}
        \mathbf{D}^{(l)} = \text{Transformer}(\{\mathbf{d}^{(l-1)}_i \odot \mathbf{W}^{(l)}_{i}\}_{i=1}^{\mid T\mid}), 
\end{aligned}
\end{equation}
$\mathbf{D}^{(l)} = \{\mathbf{d}_i^{(l)}\}_{i=1}^{|T|}$ is the integrated proposal representation in layer-$l$. $\mathbf{d}_i^{(l-1)} \in \mathbb{R}^{h}$ is the type embedding of documents in type $i$ from the previous layers of Document-level Transformer output or from a random initialized vector. The $\odot$  is a fusion operation. Inspired by ViT~\cite{dosovitskiy2020image} and TNT~\cite{han2021transformer}, we set the $\odot$ operation as:
\begin{itemize}
    \item [(1)] A \textit{Vectorization Operation} on $\mathbf{W}^{(l)}_i$.
    \item [(2)] A \textit{Fully-connected Layer} to transform the $\mathbf{W}^{(l)}_i$ from dimension $nh$ to dimension $h$. 
    \item [(3)] An \textit{Element-wise Add} with the vector.
\end{itemize}


\subsection{Selective Interpolation for Enhancing the Fairness.}
As stated in the previous section, there are only a few samples that pertain to interdisciplinary research. 
Training exclusively on the provided dataset may cause the model to overfit, and thus, it tends to infer a non-interdisciplinary path for interdisciplinary research. 
This scenario could lead to an unfair peer reviewer assignment, as the model's ability to accurately recognize and classify interdisciplinary research projects would be compromised. 
Inspired by the effectiveness of MixUp in addressing label skew and data augmentation, as demonstrated in the literature~\cite{kwon2023explainability,xiao2023traceable,xiao2023beyond,xiao2024traceable,wang2024reinforcement}, we propose a strategy of selective interpolation to resolve this issue. 
These techniques, when applied judiciously, can improve the robustness and performance of our model, particularly in cases where the dataset contains limited samples of interdisciplinary research.

\smallskip
\noindent\textbf{How to select high-quality candidate samples.}
In each training step, given a sample $\{D_i, \mathbf{L}_i\}$ and $n$ batch data $\{D_j, \mathbf{L}_j\}_{j=1}^n$, we can generate a non-parameter score from their belonging discipline, defined as:
\begin{equation}
s_{i\to j} = \frac{dis (\mathbf{L}_i, \mathbf{L}_j)}{p_{j} \cdot p_{i,j} + \epsilon},
\end{equation}
where $s_{i\to j}$ is the selective score. $dis(\cdot, \cdot)$ is used to measure the distance between two label paths. In this paper, we set the distance function as the edit distance. $p_{j}$ represents the probability of the label path $\mathbf{L}_j$ appearing in the entire dataset. $p_{i,j}$ is the conditional co-occurrence probability of $\mathbf{L}_j$ and $\mathbf{L}_i$. $\epsilon$ is a small value to avoid zero-division.
By this function, a candidate sample with a smaller share in the dataset (i.e., lower $p_{j}$), less probability to intersect with discipline $\mathbf{L}_i$ (i.e., lower $p_{i,j}$), and more distinct compared with $\mathbf{L}_i$ (i.e., higher $dis (\mathbf{L}_i, \mathbf{L}_j)$) would be more likely to be chosen for interpolation (i.e., higher $s_{i\to j}$). We feed the score of this candidate sample into a softmax function to obtain the selective probability.
\begin{equation}
    p_{i\to j} = \frac{exp(s_{i\to j})}{\sum_{k\in[1,n],k\neq i}exp\left(s_{i\to k}\right)}, 
\end{equation}
where, $p_{i\to j}$ represents the probability of selecting sample $j$ given sample $i$. By calculating this probability for each sample, we can construct a probability distribution. This distribution can then be utilized to select a sample for interpolation.

\smallskip

\noindent\textbf{MixUp the samples to even the distribution.}
We deploy a `Word-level MixUp' for semantic information interpolation.
The MixUp is a regularization technique for improving the model generalization. Most of the MixUp methods behave as a data pre-processing or a data augmentation for creating the pseudo features and the pseudo labels, so that an Empirical Risk Minimization task can be converted to a Vicinal Risk Minimization task. In the original MixUp method~\cite{Zhang2018mix}, those mixing processes of feature pair and label pair are done by using a mixing factor $\lambda$ which is sampled from a beta distribution with a hyperparameter $\alpha$:
\begin{equation}
\begin{aligned}
    f_x(x_i,x_j)&=\lambda x_i + (1-\lambda)x_j, \\
    f_y(y_i,y_j)&=\lambda y_i + (1-\lambda)y_j, \\
    \text{where}\: \lambda \sim &~ \text{Beta}(\alpha, \alpha),
    \label{mix}
\end{aligned}
\end{equation}
where the $f_x(\cdot)$ and $f_y(\cdot)$ are the original MixUp functions for features and labels. 
We adopt MixUp before the input of the Word-level Transformer. Specifically, given two type-$t$ text features $\mathbf{W}^i_t$ and $\mathbf{W}^j_t$ from $D_i$ and  $D_j$, respectively, we form the mixed feature using the MixUp function $f_x^m(\cdot)$ the Equation~\ref{mix}:
\begin{equation}
    \tilde{\mathbf{W}}_{t} = f_x(\mathbf{W}^i_t, \mathbf{W}^j_t).
\end{equation}
After that, we feed the $\tilde{\mathbf{W}}_{t}$ into the Word-level Transformer and the Document-level Transformer to obtain the pseudo document representation $\tilde{\mathbf{D}}$. 


\subsection{Topic Inference upon Hierarchical Discipline Structure}

\noindent\textbf{Why adaptively aggregate semantic information matters.}
The inference of the discipline path to which a research proposal belongs is a complex expert decision-making process in reality. For a coarse-grained discipline classification, such as distinguishing whether it belongs to Mathematical Sciences or Chemical Sciences, experts can accurately classify it simply by reading the title. However, for fine-grained inference or determining the appropriate discipline granularity, experts may need to read longer texts, such as abstracts, further. Inspired by this phenomenon, we propose an adaptive text semantic aggregation method based on predicted historical paths.
By considering the historical paths, the model can potentially learn from past experiences and patterns to make more informed decisions and generate more coherent responses or outputs. It can take into account the context provided by the previous steps to shape its understanding and reasoning, thereby improving its performance at current level.

\smallskip
\noindent\textbf{Utilizing the historical prediction.}
Suppose the model steps into level-$k$, which means it has inferred the previous $k-1$ selections, denoted by $\textbf{L}_{<k}$. To jointly model the predictive result and adaptively aggregate the separately extracted semantic information, we first look up each predicted label within each step set from a random-initialized label embedding, then Readout each step's representation:
\begin{equation}
    e_i = Readout(\textbf{H}, L_i),
\end{equation}
where $\textbf{H}$ is a random-initialized label embedding, $L_i \in \textbf{L}_{<k}$ is the predicted label set in step-$i$. The Readout operation will map each label in $L_i$ into a hidden state via $\textbf{H}$, then pool them into a vector $e_i$. In this paper, we set the pooling operation in Readout as mean pooling. Accordingly, we can mix the label embedding by using the same $\lambda$ from Equation~\ref{mix} by given the historical prediction $L^a_{i}$ and $L^b_{i}$:
\begin{equation}
\tilde{e}_i = \lambda e_i^a + (1-\lambda) e_i^b,
\end{equation}
Then, the representation of interpolated label embedding sequence can be defined as $\tilde{E}_{<k} = [\tilde{e}_i]^{k-1}_0$.

The obtained $\tilde{E}_{<k}$ will then sum with the positional encoding for preserving the prediction order. Then it will feed into a self-attention network, allowing for the propagation of information across each step:
\begin{equation}
\tilde{\textbf{E}}_{<k} = MultiHead(\tilde{E}_{<k}, \tilde{E}_{<k}, \tilde{E}_{<k}),
\end{equation}
where $MultiHead(\cdot)$ is the multi-head attention function to adaptively aggregate the context information across every step.

\smallskip
\noindent\textbf{Cross-attention based information aggregation.} 
We adopt a cross-attention mechanism to adaptively aggregate the semantic information of each document by the historical prediction results. Specifically, we use the historical representation matrix $\hat{E}$ to query the semantic information, thus adaptively aggregating them according to the prediction progress:
\begin{equation}
\begin{aligned}
    \tilde{S}_{<k} = MultiHead(\tilde{\textbf{E}}_{<k}, \tilde{\mathbf{D}}, \tilde{\mathbf{D}}),
    \label{1}
\end{aligned}
\end{equation}
where the $\tilde{S}_{<k} \in \mathbb{R}^{k\times x}$ holds the critical semantic information for the current step's prediction. We take the last element $\tilde{S}_{k-1}$ from $\tilde{S}_{<k}$ to represent the current prediction state due to it being closest to the step-$k$.


\subsection{Optimize the whole pipeline}
From the previous sections, we obtained the fusion information $\tilde{S}_{k-1}$. For a level-wise prediction, we feed the fused feature matrix $\tilde{S}_{k-1}$ into a \textit{Pooling Layer}, a \textit{Fully-connected Layer}, and a \textit{Sigmoid Layer} to generate each label's probability for $k$-th level-wise label prediction. In our paper, we set this \textit{Pooling Layer} as directly taking the last vector of $\tilde{\mathbf{S}}_{k}$. The formal definition is:
\begin{equation}
    \tilde{y}_{k} = Sigmoid(FC_k(Pooling(\tilde{S}_{k-1}))),
\end{equation}
where the $\tilde{y}_{k}$ is the probability (i.e., $P(L_k\vert D,G,\gamma,L_{<k}; \Theta)$ in Equation~\ref{propability}) for each discipline in $k$-th level of hierarchical discipline structure $\gamma$. It is worth noting that we add a $l_{stop}$ token in each step to determine the prediction state. 
Thus, the $FC_k(\cdot)$ denotes a level-specific feed-forward network with a ReLU activation function to project the input to a $\vert C_k\vert+1$ length vector, where represents the label number of level-$k$ with $l_{stop}$ token. 
After the $Sigmoid(\cdot)$, the final output $\tilde{y}_k$ is the probability of $k$-th level's labels. 
By dynamically deciding when to halt the prediction process using the $l_{stop}$ token, the model effectively tailors its predictions to the specificity of the information available for each research proposal, thereby enhancing the accuracy and applicability of the hierarchical classification.

For step-$k$ of training progress, given the selected pair $A^a$ and $A^b$, the level-wise loss function is defined as:
\begin{equation}\label{mix_objective_function}
\begin{aligned}
        \tilde{\mathcal{L}}_k(\Theta) &= \sum^{\mid C_k\mid + 1}_{i=1} \left(\tilde{Y}_k(i) \log(\tilde{y}_k^i) + (1 - \tilde{Y}_k(i)) \log(1 - \tilde{y}_k^i) \right),\\
    &\tilde{Y}_k(i) = \begin{cases}
    0 : l_i \not \in L^a_k \cup L^b_k
     \\ \lambda : l_i \in L^a_k, l_i \not\in L^b_k
     \\ 1-\lambda : l_i \in L^b_k, l_i \not\in L^a_k
     \\ 1: l_i \in L^a_k \cap L^b_k
    \end{cases},
\end{aligned}
\end{equation} 
where the $\tilde{y}_k^i$ is the $i$-th label's probability with selective interpolation. Thus, the overall loss function  $\tilde{\mathcal{L}}(\Theta)$ is formulated as:
\begin{equation}\label{overall_mix_loss}
    \tilde{\mathcal{L}}(\Theta) = \sum\tilde{\mathcal{L}}_k(\Theta).\\
\end{equation} 
During the training process with \model, our target is to optimize the objective function $\tilde{\mathcal{L}}(\Theta)$.

To infer the topic path of the provided research proposal, we can follow the pipeline in the bottom side of Figure~\ref{model_overview} and iteratively generate the label set sequence step by step.

%% file: experiment_setting.tex

\begin{table}
\centering
\caption{The discipline and sub-discipline numbers on each level of hierarchical discipline structure. }
\label{disciplines}
\resizebox{0.6\linewidth}{!}{%
\begin{tabular}{cccccc}
\toprule
Prefix & Major Discipline Name              & Total & $\mid C_2\mid$ & $\mid C_3\mid$ & $\mid C_4\mid$ \\ \midrule
A      & Mathematical Sciences              & 318          & 6       & 57      & 255     \\
B      & Chemical Sciences                  & 392          & 8       & 59      & 325     \\
C      & Life Sciences                      & 801          & 21      & 162     & 618     \\
D      & Earth Sciences                     & 166          & 7       & 94      & 65      \\
E      & Engineering and Materials Sciences & 138          & 13      & 118     & 7       \\
F      & Information Sciences               & 100          & 7       & 88      & 5       \\
G      & Management Sciences                & 107          & 4       & 57      & 46      \\
H      & Medicine Sciences                  & 456          & 29      & 427     & 0       \\ 
- & Total Disciplines & 2478 & 95 & 1062 & 1321
\\\bottomrule
\end{tabular}%
}
\end{table}

\smallskip
\noindent\textbf{Preprocessing}:
In data processing, we filter the incomplete records and group the documents in each research proposal by their types. 
We choose four parts of a research proposal as the textual data: 1) Title, 2) Keywords, 3) Abstract, and 4) Research Field. 
The Abstract part is long-text, and the average length is 100. 
The rest three documents are short text. 
All those documents are critical when experts judge where the proposal belongs. We removed all the punctuation and padded the length of each text to 100. 
On the label side, each label was provided by the fund agency, which can be treated as ground truth. We further group the labels from each sample by their level and add a stop token to the end of each label set sequence. 
As shown in Table \ref{disciplines}, when the level goes deeper, the classification becomes harder since there are more disciplines.

\begin{table}
\centering
\caption{The Details of Datasets.}
\label{dataset}
\resizebox{0.40\linewidth}{!}{%
\begin{tabular}{cccc}
\toprule
Dataset                                             & \textit{RP} & \textit{RP-IR} \\ \midrule
\multicolumn{1}{l}{Total Proposal Number}                   & 280683   &  20632     \\
\multicolumn{1}{l}{\#Avg. Labels Length} & 2.393  & 3.397
  \\
\multicolumn{1}{l}{\#Avg. Labels Num in Level-1} & 1.073  & 2.000     \\
\multicolumn{1}{l}{\#Avg. Labels Num in Level-2} & 1.197  & 2.000     \\
\multicolumn{1}{l}{\#Avg. Labels Num in Level-3} & 1.364 & 1.985     \\
\multicolumn{1}{l}{\#Avg. Labels Num in Level-4} & 0.477 & 0.808     \\
\bottomrule
\end{tabular}%
}
\vspace{-0.4cm}
\end{table}

\smallskip
\noindent\textbf{Dataset construction and description}:
We collect research proposals written by the scientists from 2020's NSFC research funding application platform, containing 280683  records with 2494 ApplyID codes. 
Among those records, 7\% are interdisciplinary proposals with two label paths, and the rest are non-interdisciplinary research. 
To compare the model performance, we construct two datasets: 
1) \textit{RP}: which contains all the records from the funding application platform. 
2) \textit{RP-IR}: which contains the IR research proposal. 
In Table \ref{dataset}, we count the lengths of each label sequence and the average number of labels in each level on these datasets. 
Generally speaking, from RP to RP-IR, its samples exhibit interdisciplinarity increasingly, and it becomes more challenging to infer correctly in the hierarchical disciplinary structure.

\smallskip
\noindent\textbf{Evaluation metrics}:\revision{
To fairly measure the model performance, we evaluate the prediction results with several widely adopted metrics \cite{metric1,metric2,xiao2021expert} in the domain of Multi-label Classification, i.e., the \textit{Precision} (P), the \textit{Recall} (R), and the \textit{F1} (F1). 
We adopt the metric Disp-Recall@1 from FairNeg~\cite{10.1145/3397271.3401177, 10.1145/3543507.3583355} to measure the group-wise fairness between IR and NIR research.
We employ the percentage degradation of the model's performance on the interdisciplinary test set compared to the overall test set to assess the unfairness caused by the interdisciplinary-non-interdisciplinary imbalance issue.}

\smallskip
\noindent\textbf{The aims of two datasets}:
The data we have collected (i.e., RP) can accurately reflect the distribution problem between interdisciplinary and non-interdisciplinary data in real-world scenarios. 
We further separated the cross-disciplinary data to create a new dataset (i.e., RP-IR). 
It's crucial to note that RP-IR would not be available before expert evaluation in real-world circumstances. 
This means that the RP-IR dataset is used solely to evaluate the model's performance on unbalanced small-scale interdisciplinary research proposals, thereby reflecting the model's ability to address issues of unfairness. 
We reiterate that the division of training, testing, and validation sets in the RP-IR dataset is entirely consistent with the RP dataset. 
This setting ensures that the model's performance comparison on the two datasets is valid.
We believe that by presenting these three sets of comparison—general model performance on RP, performance on imbalanced classes in RP-IR, and fairness via Disp-Recall—we can offer a comprehensive view of the model’s capabilities from multiple perspectives.
For each experiment, we conduct a 5-fold cross-validation evaluation. 

\smallskip
\noindent\textbf{Baseline methods}:\revision{
We compared our model with seven Text Classification (TC) methods including \textbf{TextCNN}~\cite{Kim2014}, \textbf{DPCNN}~\cite{dpcnn},  \textbf{FastText}~\cite{fasttext1},  \textbf{TextRNN}~\cite{textrnn},  \textbf{TextRNN-Attn}~\cite{textrnnattn},  \textbf{TextRCNN}~\cite{textrcnn},  \textbf{Trasnformer}~\cite{vaswani2017attention}
and three state-of-the-art HMC approaches including  \textbf{HMCN-F},  \textbf{HMCN-R}~\cite{wehrmann2018hierarchical}, and  \textbf{HARNN}~\cite{huang2019hierarchical}. 
We select four large language models, including \textbf{GPT-3.5}\footnote{\url{https://api.openai.com/v1/embeddings}}, \textbf{Llama-7b}\footnote{\url{https://huggingface.co/meta-llama/Llama-2-7b}}, \textbf{Llama-13b}\footnote{\url{https://huggingface.co/meta-llama/Llama-2-13b}}~\cite{touvron2023llama}, and \textbf{Vicuna-13b}\footnote{\url{https://huggingface.co/lmsys/vicuna-13b-v1.5}}~\cite{vicuna2023} as the text encoder, and then equipped with fully connected layers as the baselines.
We also choose six Hierarchical Multi-label Text Classification (HMTC) methods including: \textbf{HE-AGCRCNN}\footnote{\url{https://github.com/RingBDStack/HE-AGCRCNN}}~\cite{peng2019hierarchical} is an attention-based capsule network is adopted to model hierarchical relationships; 
\textbf{HiLAP}\footnote{\url{https://github.com/morningmoni/HiLAP}}~\cite{mao2019hierarchical} use strong logical constraints to reduce the number of categories for a single prediction through hierarchical dependencies; 
two variants of \textbf{HiAGM}\footnote{\url{https://github.com/Alibaba-NLP/HiAGM}}~\cite{zhou2020hierarchy} (i.e., HiAGM-LA and HiAGM-TP) that adopt pre-acquired embedded representation of hierarchical relations to assist the classification process.
\textbf{LDSGM}\footnote{\url{https://github.com/nlpersECJTU/LDSGM}}~\cite{wu2022label} is a transformer-based generative hierarchical multilabel classification method;
\textbf{HIRPCN}~\cite{xiao2023hierarchical} is a state-of-the-art transformer-based hierarchical multi-label research proposal topic inference model.
To comprehensively assess the effectiveness of MixUp, we design three variants of \model, including \textit{Word-level CutMix} (which adopts CutMix~\cite{Yun2019} to cut and concatenate two candidate samples before the hierarchical transformer), \textit{Manifold MixUp} (which applies the same MixUp strategy but to a random layer of the word-level transformer), and \textit{Document-level MixUp} (which uses the same MixUp strategy but after the hierarchical transformer, directly applying it to the generated document representation). 
We also posted four ablation models of \model\ as baselines, including \textit{- Selective Strategy} (adopt a random-based MixUp rather than Selective Interpolation),  \textit{- Selective Interpolation} (remove the interpolation entirely),  \textit{- Hierarchical Transformer} (replace the hierarchical transformer with vanilla transformer), and \textit{- Level-wise Prediction} (replace the level-wise prediction with multi-class classification head). }

%% file: experiment.tex
\begin{table}[!ht]
\centering
\caption{\revision{The overall model performance comparison. The best results are highlighted in \textbf{bold} (the higher, the better). We also post the degeneration percentage of model performance and the Disp-Recall, which could measure the fairness issue (the lower, the better).}}\label{t1}
\resizebox{0.95\textwidth}{!}{%
\setlength{\tabcolsep}{5mm}{
\begin{tabular}{lccccccc} 
\toprule
\multirow{2}{*}{\textbf{Model Name}} & \multicolumn{3}{c}{\textbf{Performance on RP Test Set}} & \multicolumn{3}{c}{\textbf{Performance on RP-IR Test Set}} & \multirow{2}{*}{\textbf{Disp-Recall}} \\
 & F1 & P & R & F1 & P & R  \\ 
\midrule
TextCNN & 0.453 & 0.455  & 0.450   &  0.373$^{\downarrow17.6\%}$&  0.495  & 0.299 & 0.262\\
DPCNN & 0.376 & 0.364  & 0.390   &  0.321$^{\downarrow14.6\%}$&  0.407  & 0.266 & 0.300\\
FastText & 0.459 & 0.451  & 0.467  &  0.381$^{\downarrow16.9\%}$&  0.493  & 0.311 & 0.244\\
TextRNN & 0.403 & 0.405  & 0.402   &  0.334$^{\downarrow17.1\%}$&  0.443  & 0.268 & 0.352\\
TextRNN-Attn & 0.409 & 0.411  & 0.408   &  0.339$^{\downarrow17.1\%}$&  0.448  & 0.272 &  0.367\\
TextRCNN & 0.423 & 0.423  & 0.423   &  0.357$^{\downarrow15.6\%}$&  0.469  & 0.288 & 0.303\\
Transformer & 0.369 & 0.356  & 0.382   &  0.310$^{\downarrow15.9\%}$&  0.394  & 0.256 & 0.248 \\
\midrule
GPT-3.5 & 0.510 & 0.497 & 0.525 & 0.434$^{\downarrow14.9\%}$ & 0.555 & 0.356 & 0.284\\
Llama-7b & 0.531 & 0.518 & 0.544 & 0.452$^{\downarrow14.8\%}$ & 0.578 & 0.372 & 0.270\\
Llama-13b & 0.528 & 0.503 & 0.555 & 0.454$^{\downarrow14.0\%}$ & 0.559 & 0.382 & 0.280\\
Vicuna-13b & 0.517 & 0.496 & 0.539 & 0.447$^{\downarrow13.5\%}$ &0.554 & 0.374 & 0.293\\
\midrule
HMCN & 0.679 & 0.656  & 0.685   &  0.541$^{\downarrow20.3\%}$&  0.683  & 0.448 & 0.226\\
HMCN-R & 0.594 & 0.593  & 0.594   &  0.469$^{\downarrow21.2\%}$&  0.612  & 0.380 & 0.238\\
HARNN & 0.686 & 0.676  & 0.696   &  0.546$^{\downarrow20.4\%}$&  0.700  & 0.448 & 0.234\\
\midrule
HE-AGCRCNN&  0.513 & 0.502 & 0.533 & 0.403$^{\downarrow21.3\%}$ & 0.532 & 0.321 & 0.281\\ 
HiLAP & 0.628 & 0.618 & 0.632 & 0.489$^{\downarrow22.1\%}$ & 0.623 & 0.401 & 0.255\\
HiAGM-LA & 0.594 & 0.593 & 0.595 & 0.476$^{\downarrow19.9\%}$ & 0.608 & 0.387 & 0.232\\
HiAGM-TP & 0.629 & 0.608 & 0.642 & 0.506$^{\downarrow19.6\%}$ & 0.626 & 0.409 & 0.187\\
LDSGM & 0.701 &  0.693 & 0.704 & 0.573$^{\downarrow18.2\%}$ & 0.715 & 0.452 & 0.145\\
HIRPCN & 0.749 & 0.726  & 0.774   &  0.615$^{\downarrow17.8\%}$&  0.763  & 0.516 & 0.118\\
\midrule
\textbf{\model}     & \textbf{0.781} &  \textbf{0.758 } & \textbf{0.805}    &   \textbf{0.704}$^{\downarrow\mathbf{9.8\%}}$& \textbf{0.827}  & \textbf{0.613}  & \textbf{0.109}\\
\bottomrule
\end{tabular}}}
\end{table}
\subsection{Overall Comparison}\label{main_exp}
\begin{figure*}[t]
\centering
\subfigure{
\includegraphics[width=7.3cm]{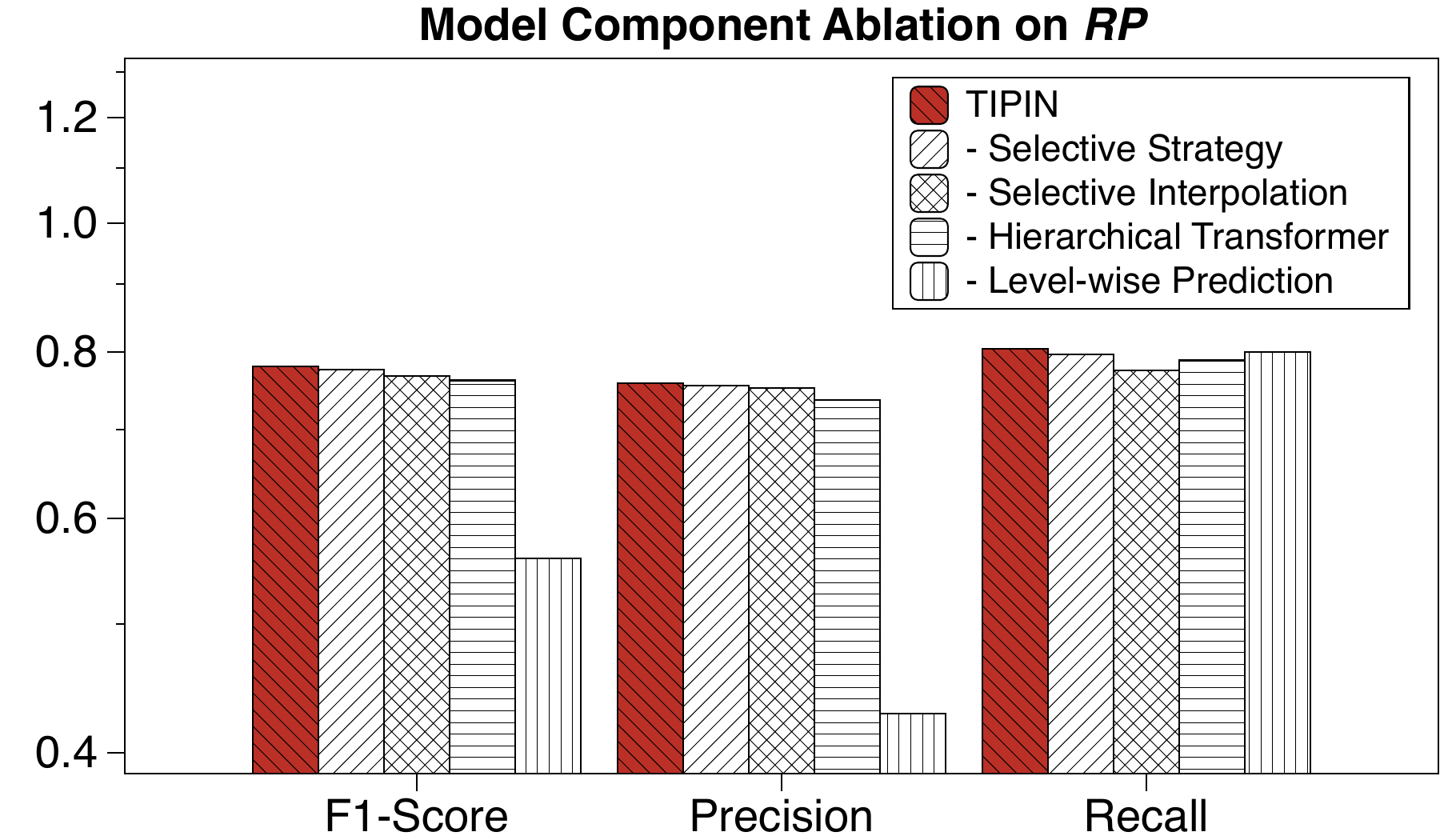}
}
\subfigure{
\includegraphics[width=7.3cm]{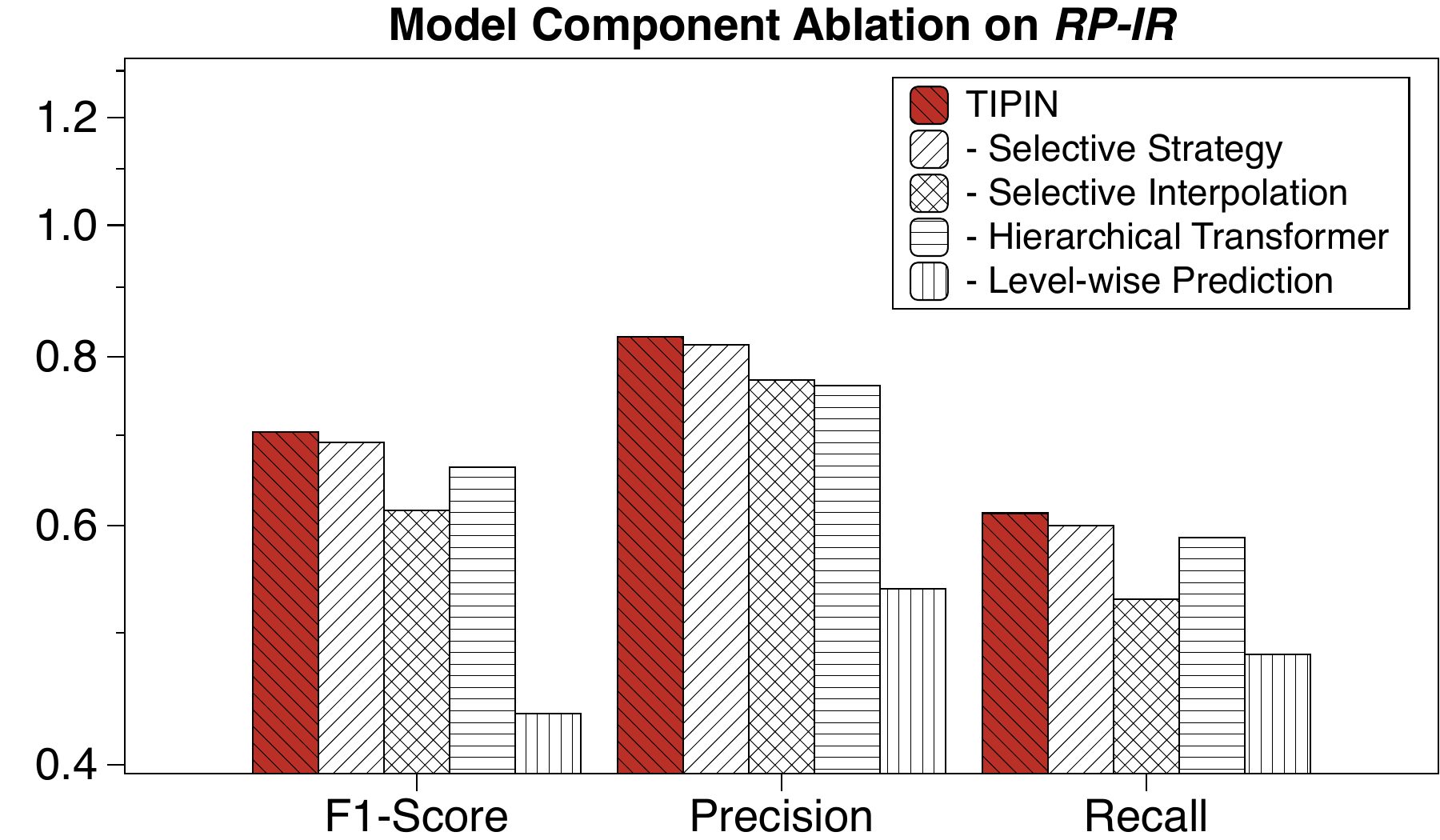}
}\\
\subfigure{
\includegraphics[width=7.3cm]{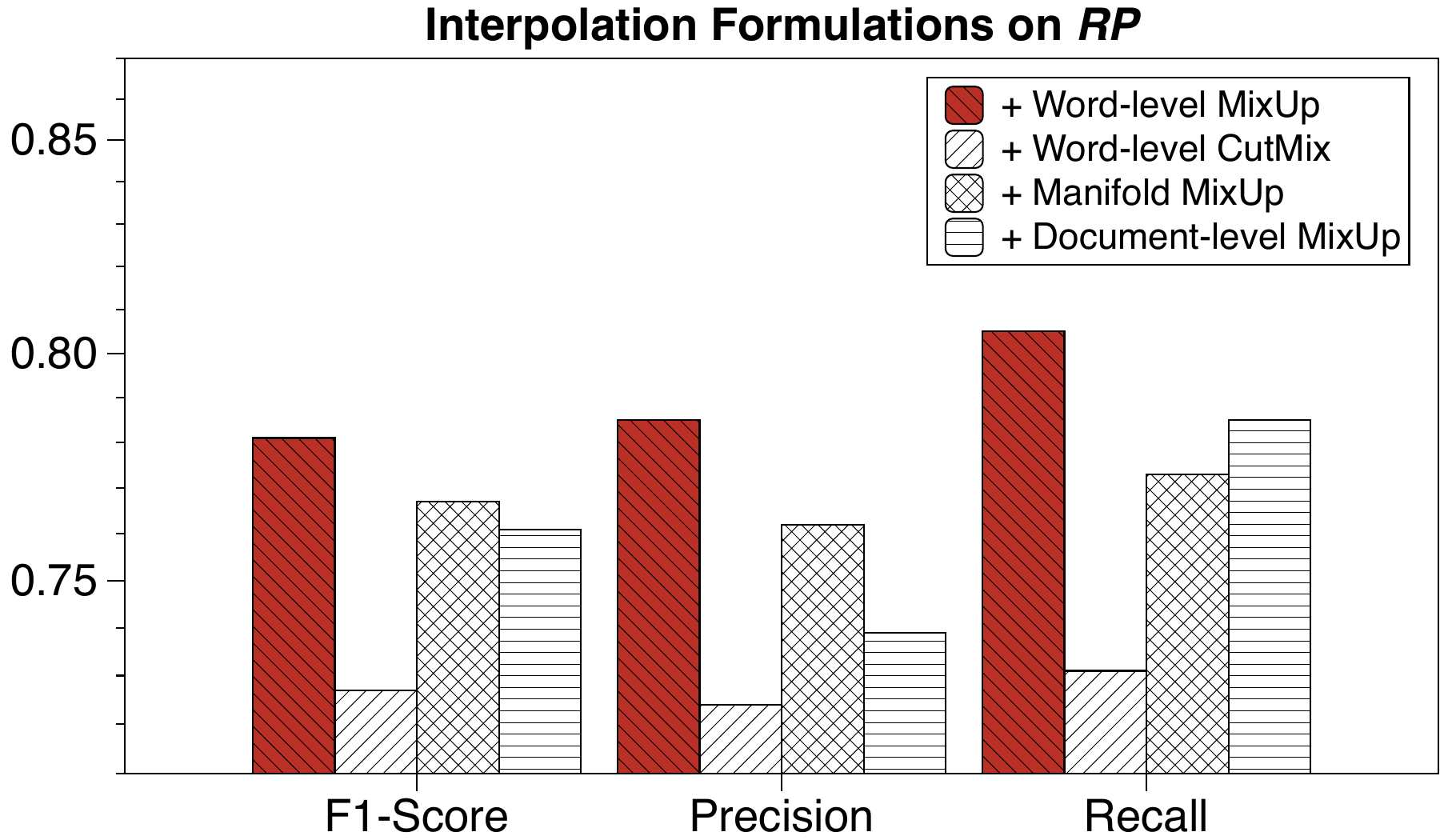}
}
\subfigure{
\includegraphics[width=7.3cm]{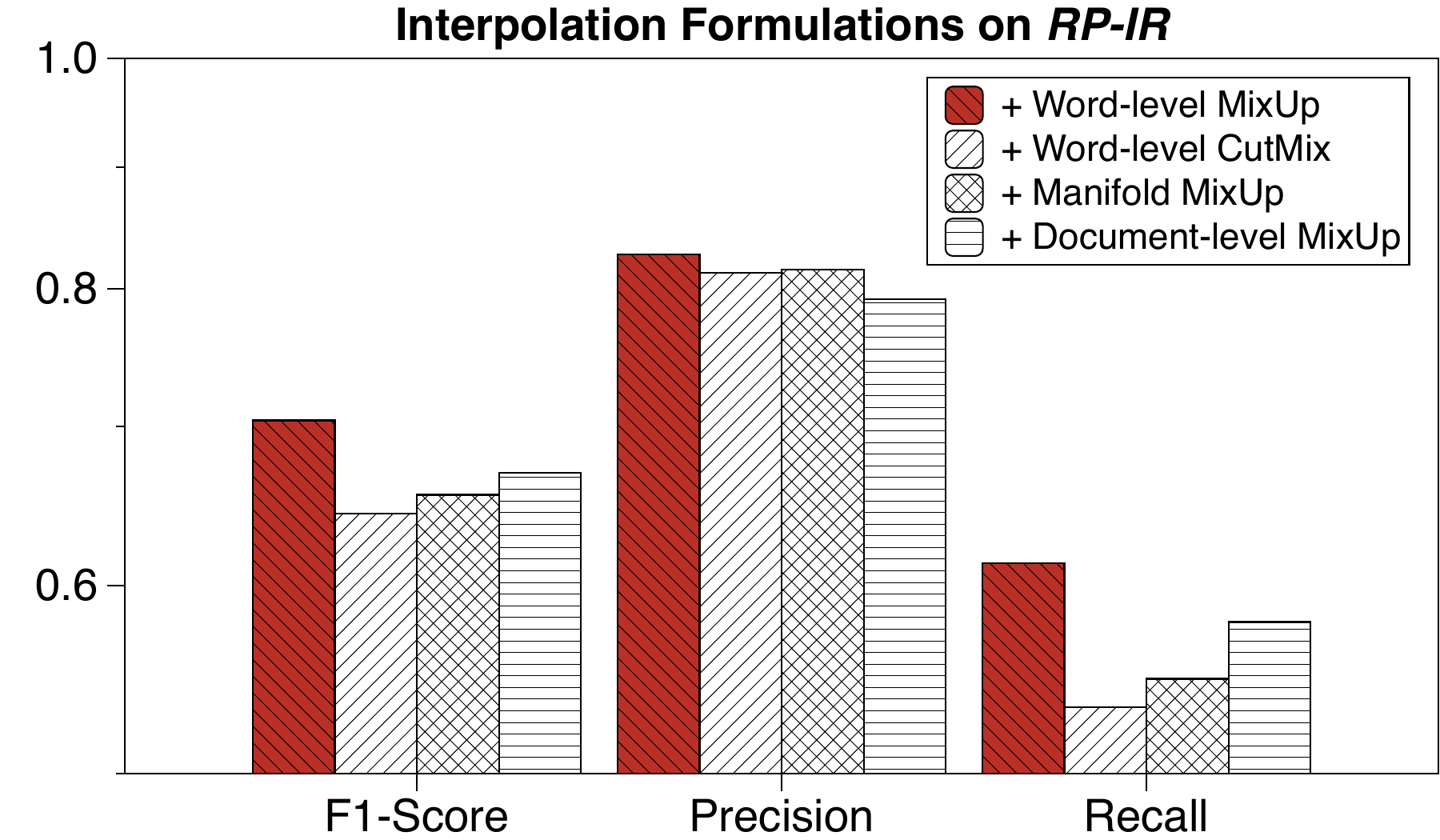}
}
\caption{The study of model component ablation and interpolation formulation impact.}
\label{figure.ablation}
\end{figure*}

\revision{
This experiment aims to answer: \textit{Can \model\ effectively imitate the fairness issue with excellent performance?}
We initially trained the baselines and \model\ on the imbalanced dataset \textit{RP}. 
Subsequently, we tested each method on the test set of \textit{RP} and \textit{RP-IR}. The results are reported in the first two figures in Figure~\ref{figure.ablation}.
Firstly, we observed that the performance of each method declined when transitioning from \textit{RP} to \textit{RP-IR}. 
This indicates that data imbalance affects model predictions irrespective of whether the model employs a flat prediction approach (i.e., TC methods) or follows a top-down fashion (i.e., HMC methods and the original method). 
Specifically, the F1 score for each TC method decreased by -14.6\% (DPCNN) to -17.6\% (TextCNN). Similar declines were observed in the HMC (-20.3\% to -21.2\%), LLMs (-13.5\% to -14.9\%), and HMTC (-17.8\% to -22.1\%). 
This performance drop underlines the severe issue caused by the data imbalance between IR and NIR data, which will result in a serious fairness issue. 
We also observed that Recall (R) declined more drastically than Precision (P), leading to a lower F1 score. 
This aligns with the statement in the introduction, where training data can cause the model to predict incomplete label paths for IR data. 
Compared to TC methods, we observed that the margin of deterioration in HMC methods was larger. We attribute this to the HMC methods' focus on designing sophisticated classifiers, making them more prone to degradation when label numbers are imbalanced. 
We can also observe that LLM baselines achieve slightly better than TC methods but worse than HMC and HMTC methods. 
The underlying driver is that although those LLM-based methods could extract semantic information well, their decoder lacked awareness of the hierarchical structure, which led to inferior performance. 
Besides, we found that HiAGM-TP has a better fair than its variant HiAGM-LA. The underlying driver is that HiAGM-TP adopted a structure encoder to aggregate the semantic information to node representation across the hierarchy structure. 
Those learned node embeddings help the model to conduct a fair prediction result. 
We could also find that LDSGM and HIRPCN have a relatively lower Disp-Recall score. 
According to the analysis between HiAGM-TP and HiAGM-LA, both of the two transformer-based methods use graph neural network-related methods to model the node representation on the label hierarchy, which could be the underlying reason that brings a better fair prediction. 
Besides, \model\ achieves the highest F1 score and Disp-Recall, indicating that the \model\ is the most effective and fair for the topic inference problem among each baseline. 
The underlying driver could be the selective interpolation strategy adopting the co-occurrence between the two topics as the score for interpolation. 
Those generated pseudo-research proposals could be much harder to classify, thus making the model more robust. 
}

\subsection{Discussion on Model Component}\revision{
This experiment aims to answer: \textit{How does each component in \model\ impact its performance?} 
In order to methodically appraise the individual contributions of our model components, we initiated an ablation study. As part of this investigation, four distinct variations of our \model\ were developed and examined. These variations, specifically `- Selective Strategy', `- Selective Interpolation', `- Hierarchical Transformer', and `- Level-wise Prediction', were detailed in the experiment setting section of our study.
The first two figures in Figure~\ref{figure.ablation} offer a comprehensive report of the performance of each of these ablation cases. A preliminary observation from the table suggests that the \model\ variant, devoid of the selective strategy (i.e., replaced by a random selection of the MixUp sample), demonstrates inferior performance compared to the standard \model. The fundamental rationale behind this observation can be attributed to the strategic selection of high-quality candidate samples by \model\ for mixing purposes, which not only balances the distribution of the dataset but also tends to amalgamate two discipline-distinct samples.
The comparison of `- Selective Strategy' and \model\ with `-Selective Interpolation' reveals superior performance on all metrics, underlining the efficiency of our interpolation component in enhancing model performance. This observation also provides important insights into addressing the fairness issue in the peer-reviewer assignment process.
Secondly, we observed that the overall performance decline of the \model\ variants (i.e., -9.8\% to -14.6\%) is lower than that of other methods and considerably better than those ablated from the selective interpolation (i.e., `- Selective Interpolation') ones (+4.5\% to +14.4\%). 
This suggests that introducing selective interpolation can mitigate the data imbalance issue during training. In addition to this, the experimental results also demonstrated improvements in the F1 score and other measurements. 
An interesting outcome from our study is the relatively poorer performance of `- Hierarchical Transformer' as compared to the \model. 
This reinforces the notion that separate modeling of each document tends to be more effective for research proposal data.
A stark contrast in performance can also be observed with `- Level-wise Prediction', which underperforms substantially compared to the other ablation cases. 
The root cause for this can be traced back to 1) the adaptive aggregation of semantic information during the inference process will be disabled. 
2) the flattening of the hierarchical label structure in this variation leads to the model having to classify the research proposal into an excess of 2,000 classes. 
This observation is consistent with the conclusion in Section~\ref{main_exp}, that the model with a weak decoder (e.g., TC and LLMs baselines) will perform inferiorly.}
%

\subsection{Discussion on Interpolation Formulation}
This experiment aims to answer: \textit{Which interpolation formulation can better address the fairness issue?} 
To conduct this experiment, we design three variants of \model\ in the experiment setting section, i.e., + World-level CutMix, + Manifold MixUp, and + Document-level MixUp. 
We follow the same setting in overall comparison and report each variant's performance in Table~\ref{t1}. 
Compared to the variant without Selective Interpolation (i.e., `- Selective Interpolation'), \model\ (` + Word-level MixUp', +4.27\%), `+ Manifold MixUp' (+2.4\%), and `+ Document-level MixUp' (+1.6\%) all achieved higher F1 scores in \textit{RP}.
This implies that all \model\ variants can enhance the categorization of interdisciplinary samples by utilizing mixed non-interdisciplinary sample data. 
Moreover, the mixed feature can also improve the model's performance for non-interdisciplinary samples (i.e., in the RP test set). 
However, we also note that `+ Word-level CutMix' performs worse than other \model\ variants. We believe that concatenating the text to mix features may lead to a significant loss of semantic information in the text classification task, thereby impacting the model's performance.
This experiment results indicate that the interpolation method in \model\ is the most effective among other variants.

\begin{figure}[!t]
\subfigure{
\includegraphics[width=7.3cm]{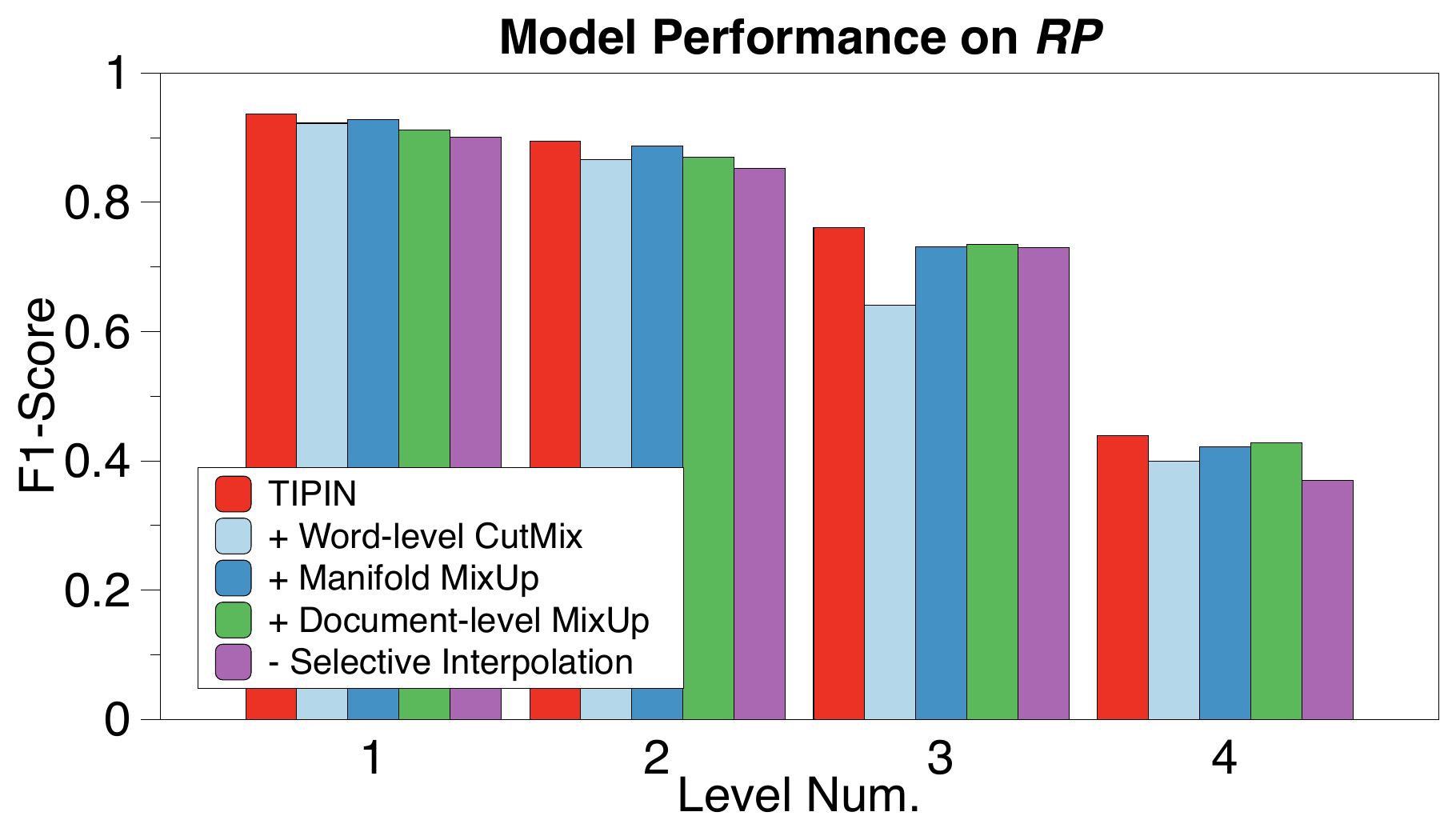}
}
\subfigure{
\includegraphics[width=7.3cm]{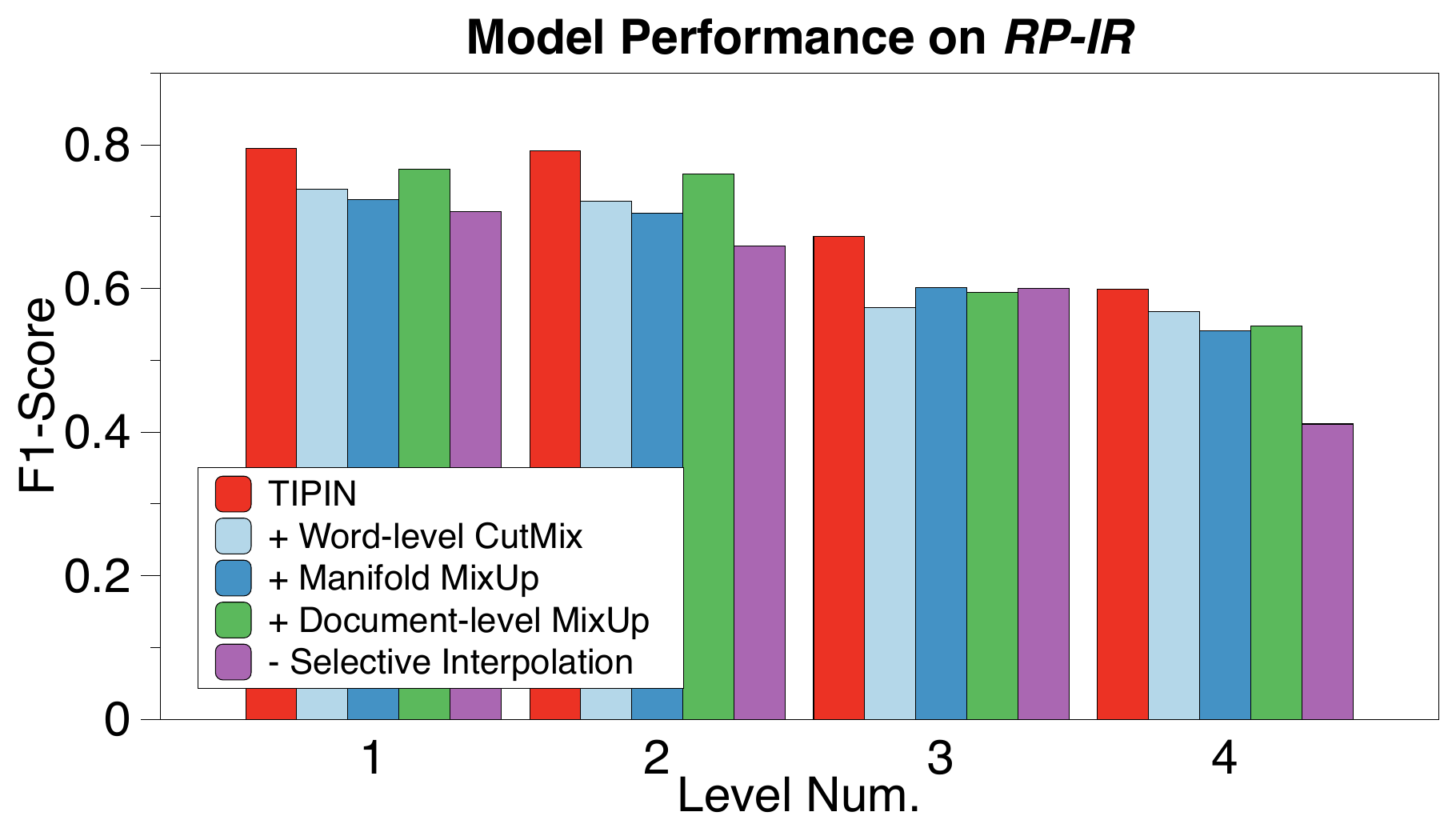}
}
\caption{Level-wise prediction results of each \model\ variants.}
\label{figure.level}
\vspace{-0.4cm}
\end{figure}

\subsection{Discussion on Level-wise Performance Comparison with Different Interpolation Formulations}
This experiment aims to answer: \textit{How do different interpolation formulations affect level-wise performance?}
We further reported the performance of level-wise prediction on the \textit{RP} and \textit{RP-IR} in Figure~\ref{figure.level} to show the advantage of our proposed interpolation strategy on each level. 
From the research funding administrators' perspective, the accuracy of the last label should be the most important because the finest-grain categorization would significantly improve the reviewer assignment. 
Based on the results in  Figure~\ref{figure.level}, we can draw the following conclusions.
Firstly, the performance of each model tended to decrease with the increase in depth of level. This phenomenon is because the number of categories increases rapidly while the depth goes deeper, making the classification task harder.
As a subsequent observation, it is noteworthy that as the level deepens, the performance of the `- Selective Interpolation' deteriorates significantly compared to other variants. The underlying causality of this can be traced to the imbalance issue that amplifies the scarcity of the target label number in the last two levels of training data while the candidate label numbers are increasing simultaneously.
However, the \model\ variants demonstrate proficient handling of this predicament, substantiating the effectiveness of our proposed strategy in enhancing model performance when dealing with imbalanced datasets.
Of all the variations, \model\ consistently achieves the most commendable performance at each level. This finding aligns with the performance metrics of \model\ as stated in Table~\ref{t1}, reinforcing the robustness and reliability of our primary model.

\begin{figure}[!h]
\centering
\subfigure{
\includegraphics[width=7.3cm]{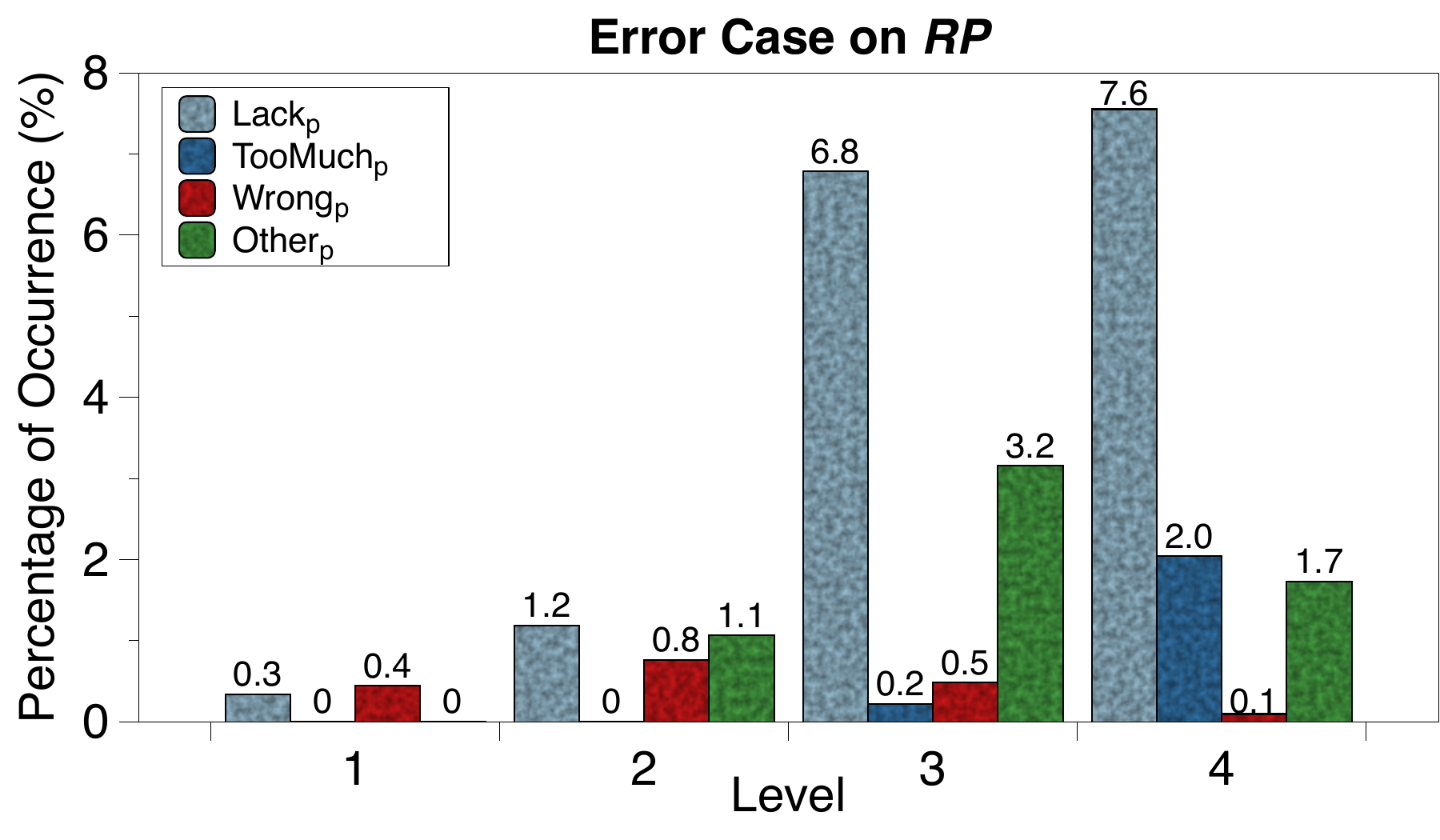}
}
\subfigure{
\includegraphics[width=7.3cm]{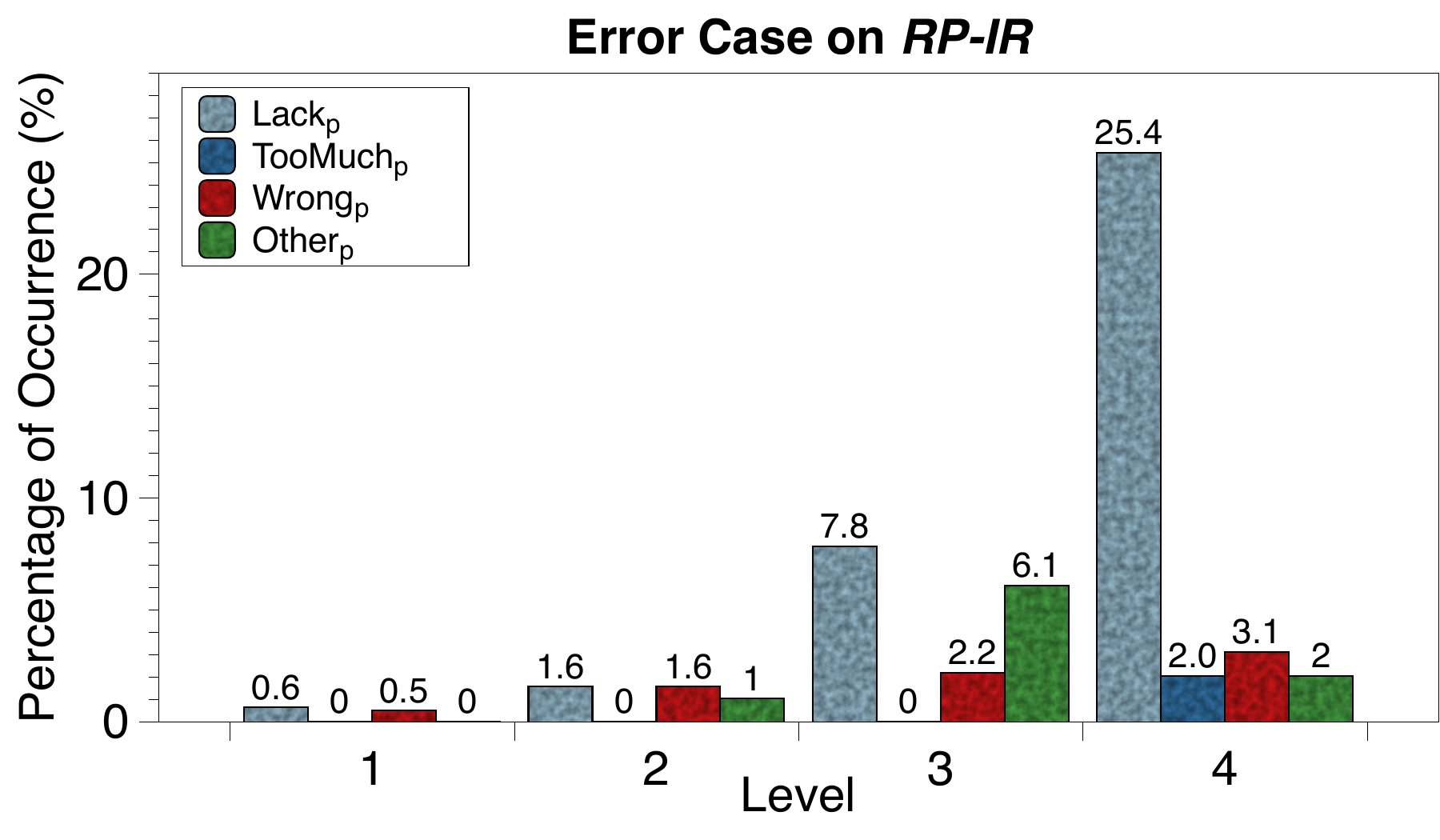}
}
\caption{The details of the wrong case occurrence rate on each dataset.}
\label{figure.wrong}
\vspace{-0.4cm}
\end{figure}

\subsection{Discussion on Error Case}
This experiment aims to answer: \textit{What is the main type of incorrect prediction?}
In this study, we quantitatively assessed the frequency of four distinct types of prediction errors within our model: \textbf{Lack}, \textbf{TooMuch}, \textbf{Wrong}, and \textbf{Other}. The \textbf{Lack} error typifies instances where the model fails to generate a prediction at the requisite granularity, halting at a more generalized level. Conversely, \textbf{TooMuch} denotes scenarios where the model overextends into an excessively detailed level despite the absence of a corresponding truth label at this granularity. \textbf{Wrong} indicates predictions that deviate from the established hierarchical structure. Lastly, \textbf{Other} encompasses all remaining error categories.
The experiment result was reported in Figure~\ref{figure.wrong}.
Firstly, our observations indicate that \textbf{Lack} is the most frequent cause of disparity between the model's predictions and the truth labels. In contrast, \textbf{TooMuch} occurred less frequently, implying a conservative approach by the model in advancing through the topic hierarchy. 
We speculate that the underlying reason is that the label has become more scarce in the last few levels, and the $l_{stop}$ is emerging more often in most training data, making the model tend to predict stop tokens.
Regarding the \textbf{Wrong} category, it constituted approximately 0.4\% to 3.1\% of errors at each level, indicating that despite the absence of an explicit constraint enforcing coherence, the model inherently learns and adheres to the underlying dependencies and hierarchical structure. 
Furthermore, our analysis revealed a progressive increase in the incidence of these errors from level-1 to level-4, mirroring the observed decline in metric performance as depicted in Figure~\ref{figure.level}. 
Notably, at level 4, the predominant error was \textbf{Lack}, suggesting that the model's tendency to halt predictions is a more prevalent issue than incorrect prediction generation. This type of error, while not ideal, is considered more tolerable in practical applications.

\begin{figure}[!h]
    \centering
    \includegraphics[width=0.40\linewidth]{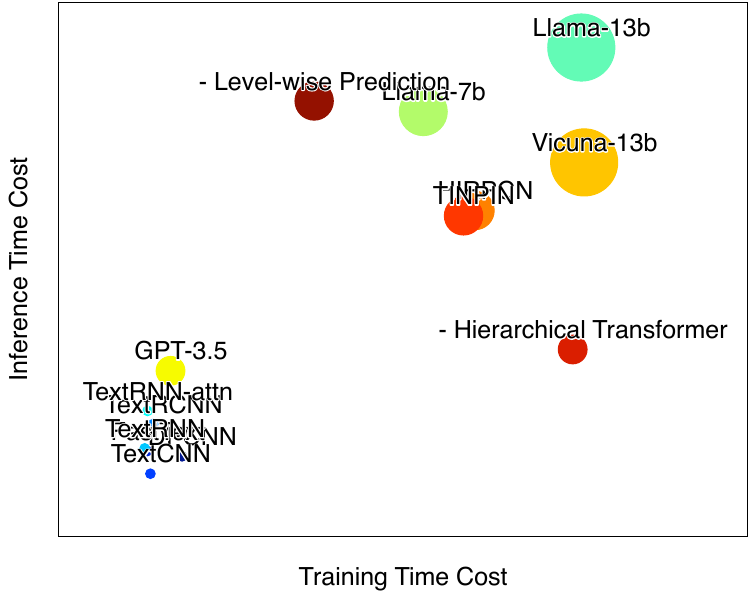}
    \caption{The illustration of the time cost and model size on selected baselines. }
    \label{fig:times}
\end{figure}

\revision{\subsection{Discussion on Training and Inference Time Cost}
This experiment is designed to answer the question:"\textit{What is the overhead of this framework?}". 
To address this research question, our study measured the training time, inference time, and the size of various models. The time cost is observed to increase from the bottom-left to the top-right, with larger dots indicating models of a larger scale. 
It was first noted that non-Transformer-based methods (i.e., most text classification (TC) baseline methods) required shorter training/inference times compared to others. 
This phenomenon is primarily attributed to their relatively lower model complexity. 
Additionally, GPT-3.5 demands less time in training and inference phases due to its output vector dimension being 1536, whereas other Large Language Models (LLMs) like Llama-7b have a dimension of 4096, and Llama13b and Vicuna-13b stand at 5120. Furthermore, we observed that Transformer-based methods such as \model\ and HIRPCN exhibit faster training and inference speeds compared to directly using large language models. 
It was also noted that compared to \model\ , its ablated Decoder version `-Level-wise Prediction' demonstrates quicker training speeds but slower inference speeds. 
Conversely, another ablated Encoder version `- Hierarchical Transformer' shows faster inference but slower training speeds. 
From this comparison with \model\ , we infer that the primary computational expenditure in \model\ is during the training of the Decoder. 
Once trained, due to the task form of hierarchical multi-label classification, the inference of labels is faster than direct global prediction.
}


\begin{figure*}[!h]
\centering
\subfigure{
\includegraphics[width=7.3cm]{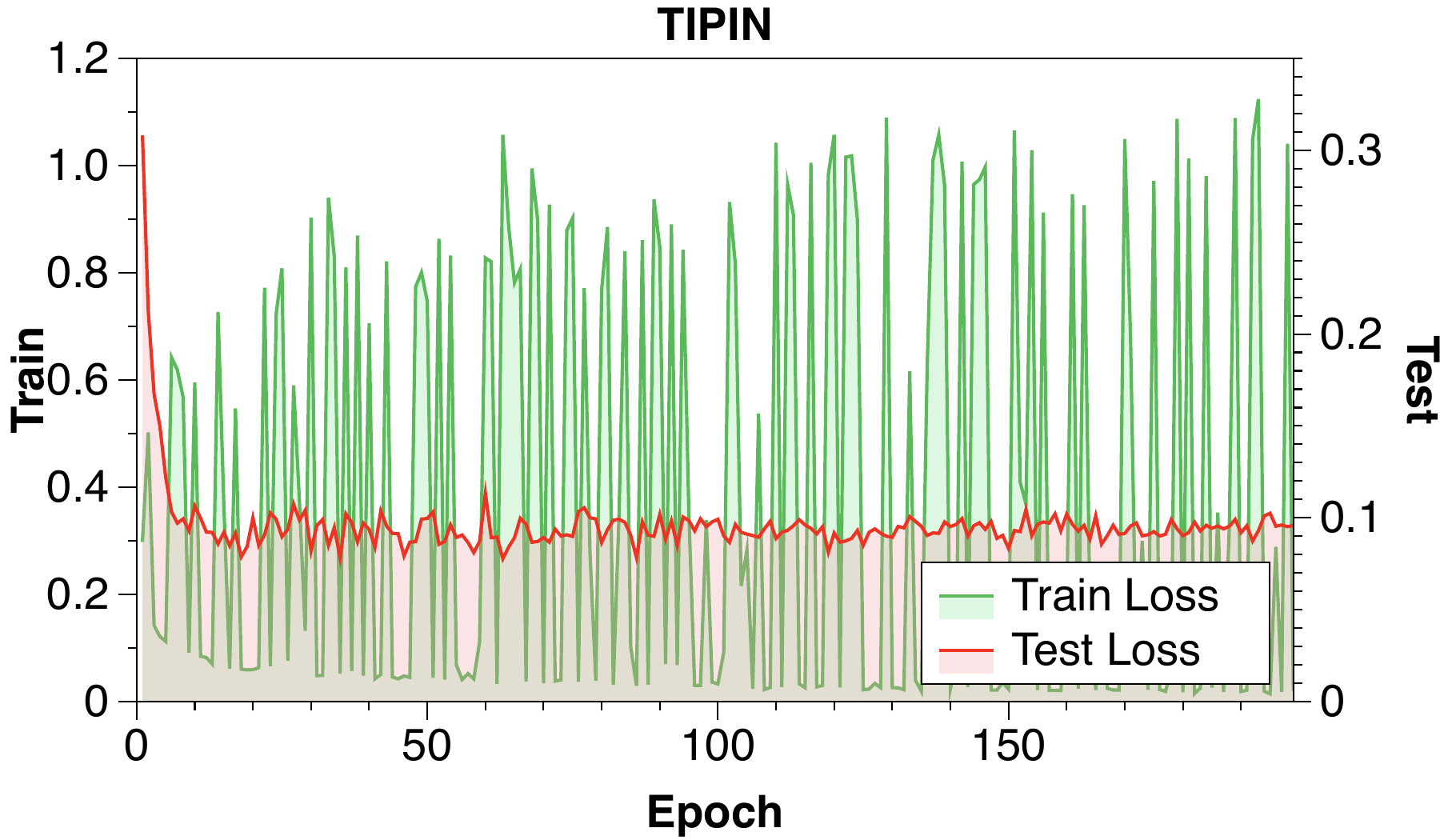}
}
\subfigure{
\includegraphics[width=7.3cm]{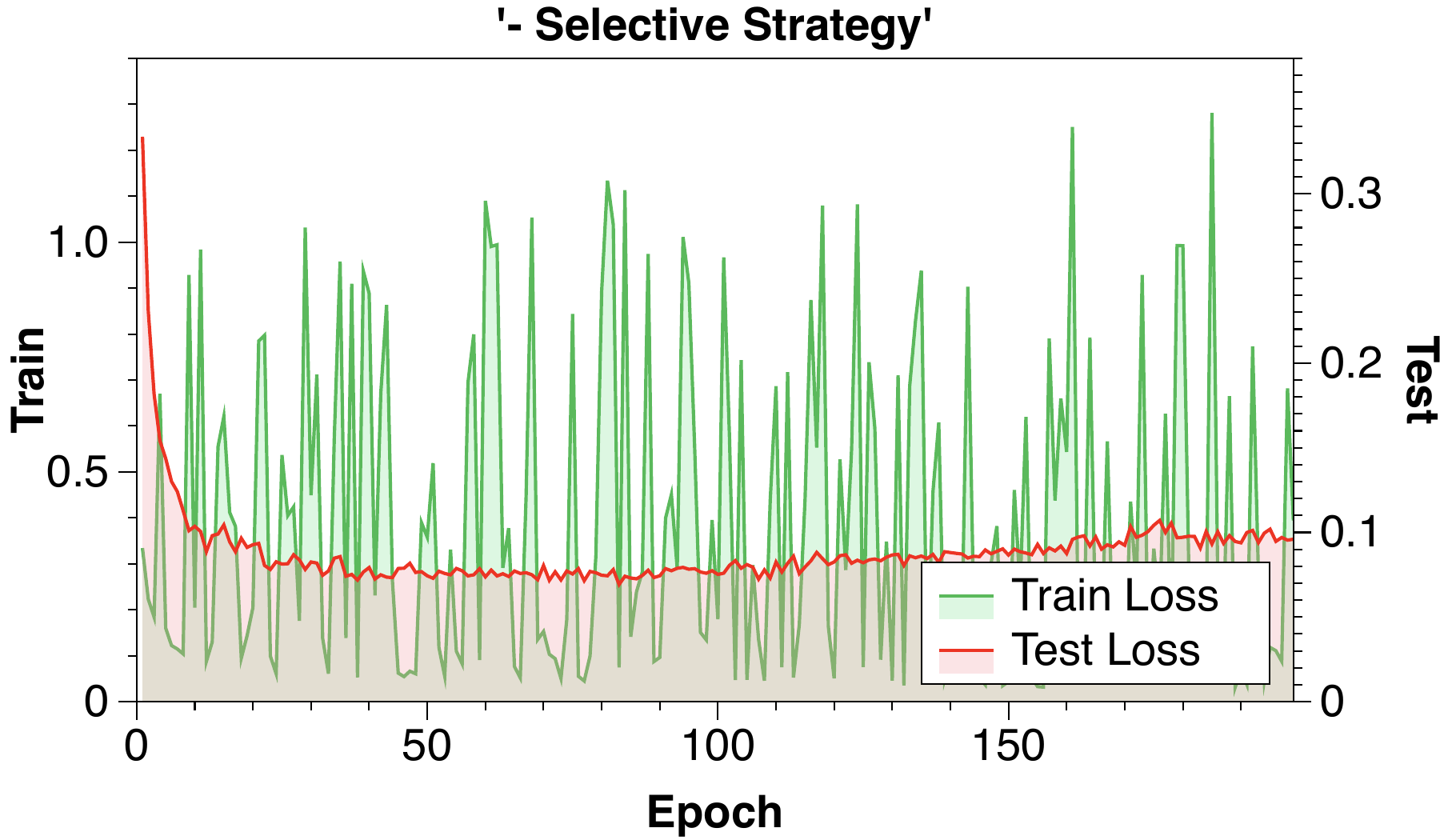}
}\\
\subfigure{
\includegraphics[width=7.3cm]{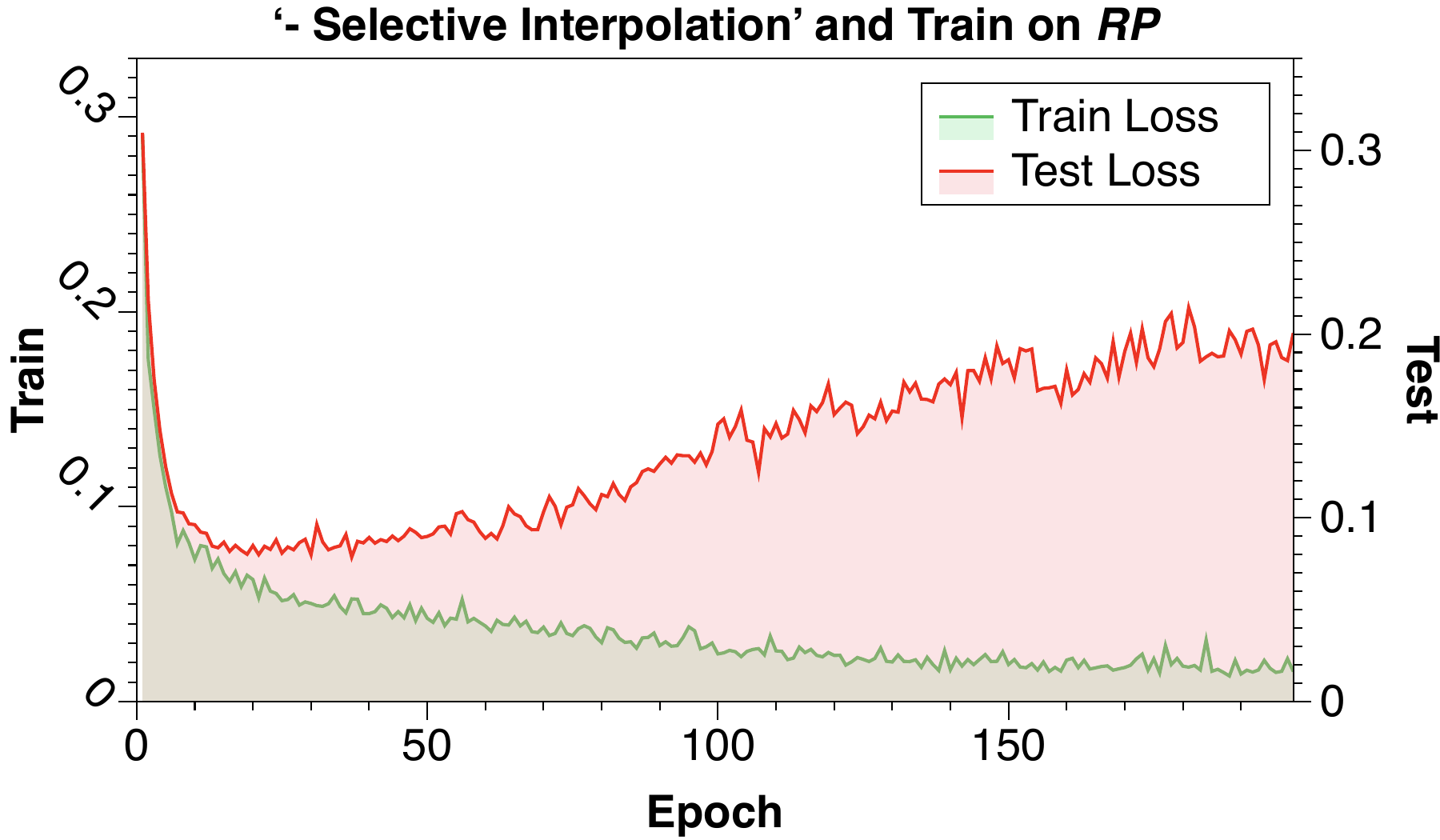}
}
\subfigure{
\includegraphics[width=7.3cm]{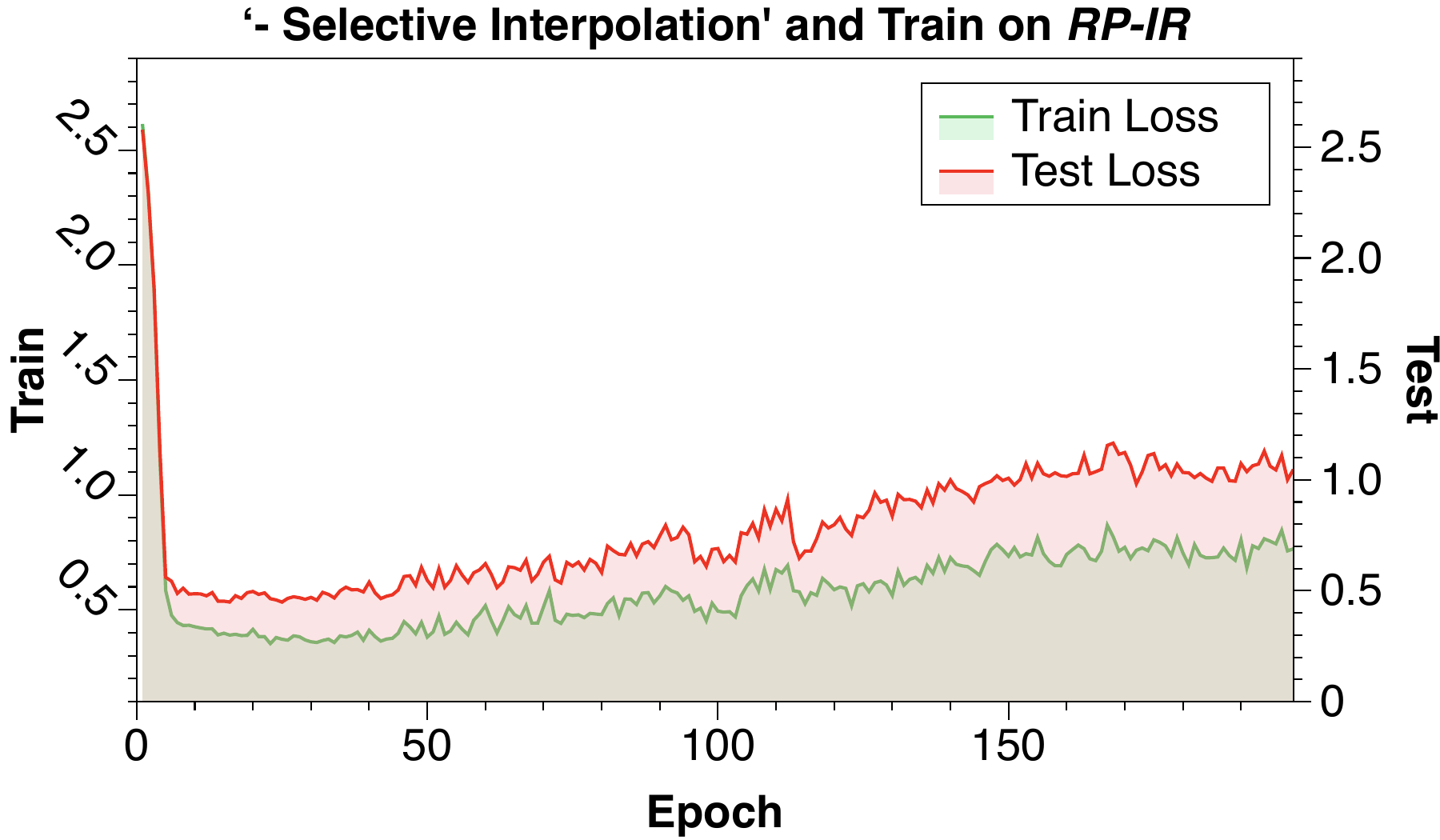}
}
\caption{The train loss and test loss on RP-IR during the training process with different selective interpolation formulations.}
\label{figure.loss}
\end{figure*}

\subsection{Discussion on Training and Testing Loss}
This experiment is designed to answer the question: \textit{"What is the impact of selective interpolation on the training process?"} To address this, we study the training and testing losses of three ablation cases of the \model\ model, as depicted in Figure~\ref{figure.loss}. The three cases include the original \model, and two variants, namely `- Selective Strategy' and `- Selective Interpolation'.

It is observable from the figure that removing the selective interpolation component can cause a significant increase in the testing loss. This is likely due to the model's propensity to misclassify interdisciplinary research proposals as non-interdisciplinary ones, thereby leading to a significant fairness issue. However, it's worth noting that the last two methods, which have the selective interpolation component removed, exhibit distinct behaviors. This could be attributed to the fact that these models are trained on the RP-IR dataset, which exclusively comprises interdisciplinary research proposals, thereby eliminating the fairness issue.
The first two figures elucidate the impact of MixUp techniques on model training. We observe drastic fluctuations in the training loss and a consistent decline in the testing loss with the introduction of mixed features. This observation underscores the role of interpolation in enhancing model generalization and training stability, especially when dealing with imbalanced datasets such as the one comprising interdisciplinary and non-interdisciplinary proposals.
On comparing the `- Selective Strategy' variant with the original \model, it is found that the selective interpolation technique leads to more drastic changes in the training loss and a more stable decrease in the testing loss on the RP-IR dataset. This can be ascribed to the fact that Selective Interpolation selects two high-quality samples for MixUp, thereby introducing greater complexity into the model training. The stable decline in the testing loss corroborates the effectiveness of this strategy in mitigating the fairness issue in the topic inference task.


\begin{figure}[!h]
\centering
\includegraphics[width=0.85\linewidth]{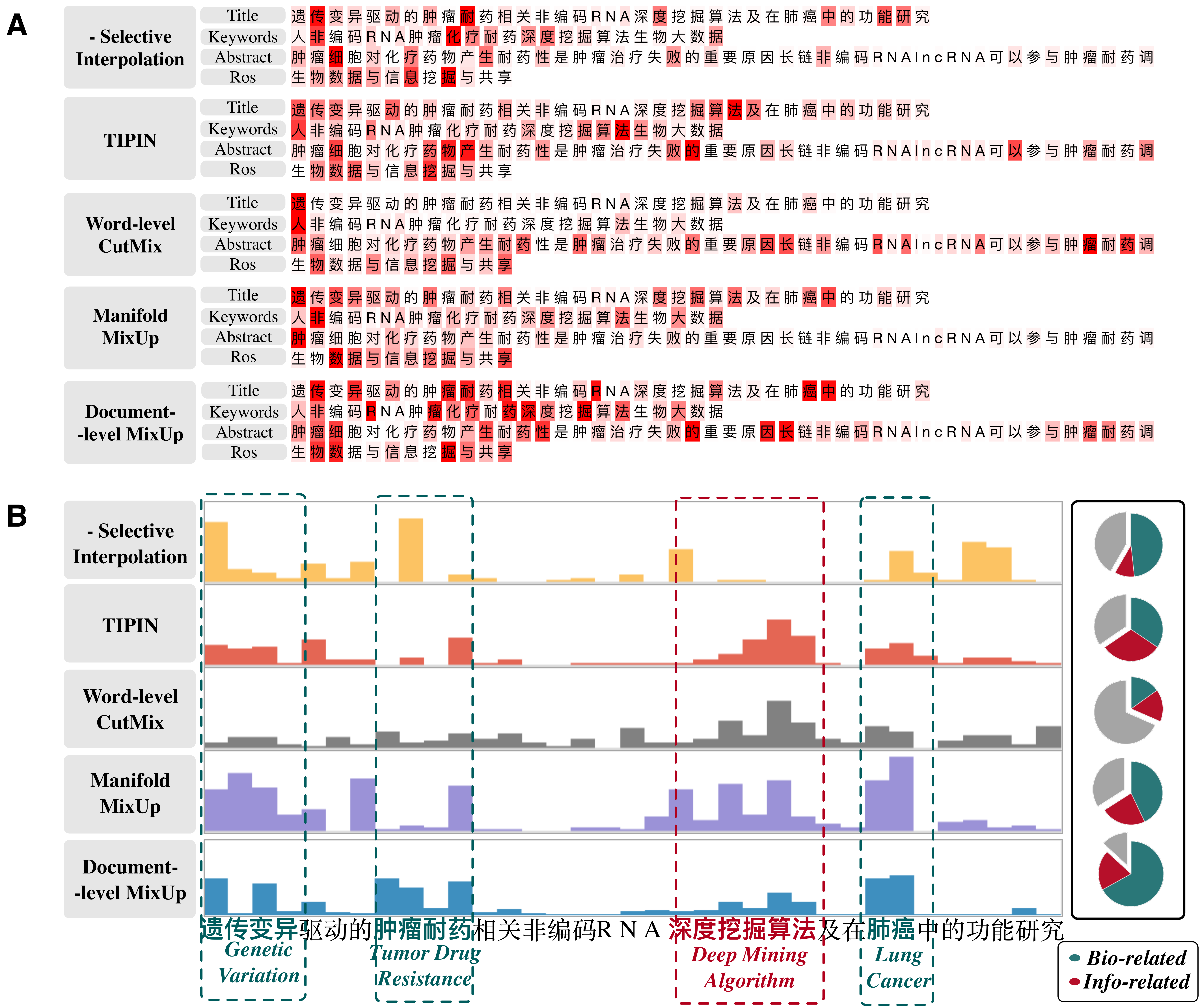}
\caption{The attention value to an example interdisciplinary research proposal with different \model\ variants. (A) The attention visualization on each component of the example research proposal. (B) The attention visualization on the \textit{Title} of example research proposal. We also highlight the keywords and the attention value proportion from different research areas in green and red colors.}
\label{figure.attn}
\vspace{-0.4cm}
\end{figure}

\subsection{Discussion on Attention Mechanism}
\begin{CJK}{UTF8}{gbsn}
This experiment aims to answer: \textit{How dose the selective interpolation affect the attention during training?}
We further discuss how the distinct interpolation methods affect the attention mechanism on the text data.
In Figure~\ref{figure.attn}, we illustrated the title of interdisciplinary research from the \textit{Life Sciences} and the \textit{Information Sciences}. 
We first colored the Information Sciences-related (Info-related) keywords {red} and the Life Sciences-related (Bio-related) keywords {green}, then fed the selected text sequence into the original model and four ablation variants. 
The middle part of the figure shows the model's attention value to this sequence, and the right side of this figure illustrates the proportion of the model's attention to the different domains.

From Figure~\ref{figure.attn} (A), we can observe that by using the selective interpolation strategy, the model could focus more on some domain-related keywords, enhancing the model performance. 
From Figure~\ref{figure.attn} (B), we can further observe that except for the `+ Word-level CutMix', all other models gained more attention on the domain-related keywords such as 
{遗传变异} (Genetic Variation, a Bio-related keyword) and {深度挖掘算法} (Deep Mining Algorithm, an Info-related keyword), proving that the selective interpolation can help the model to capture the critical part of the input research proposal. 
We speculate that the CutMix-based H-MixUp performed worse due to its concatenating of text, undermining the sequence's positional information and dragging the self-attention mechanism, which is consistent with the result in Table~\ref{t1}.
On top of that, we found that the MixUp-based selective interpolation will enhance the model to capture the critical words more efficiently, e.g., the proportion of colored shading in \model\ is more significant than the `- Selective Interpolation'. We speculate that the MixUp technique will bring more combinations of the text and the discipline information so the model can better learn and capture the unique critical part of the research proposal.
Besides this, compared to the other selective interpolation strategies, the ones on the word-level (i.e., \model, and `+ Word-level CutMix') place more balanced attention on the two domain-related keywords than the `+ Document-level MixUp' or the `+ Manifold MixUp'. We conclude that the former level of feature mix in Transformer training will help the model get more balanced attention. 
Based on the above discussion, we summarize that \model\ gains the highest F1 scores due to its most balanced attention to the text sequence and better capture of domain-related keywords.
\end{CJK}

\begin{figure*}[!h]
\centering
\subfigure{
\includegraphics[width=7.3cm]{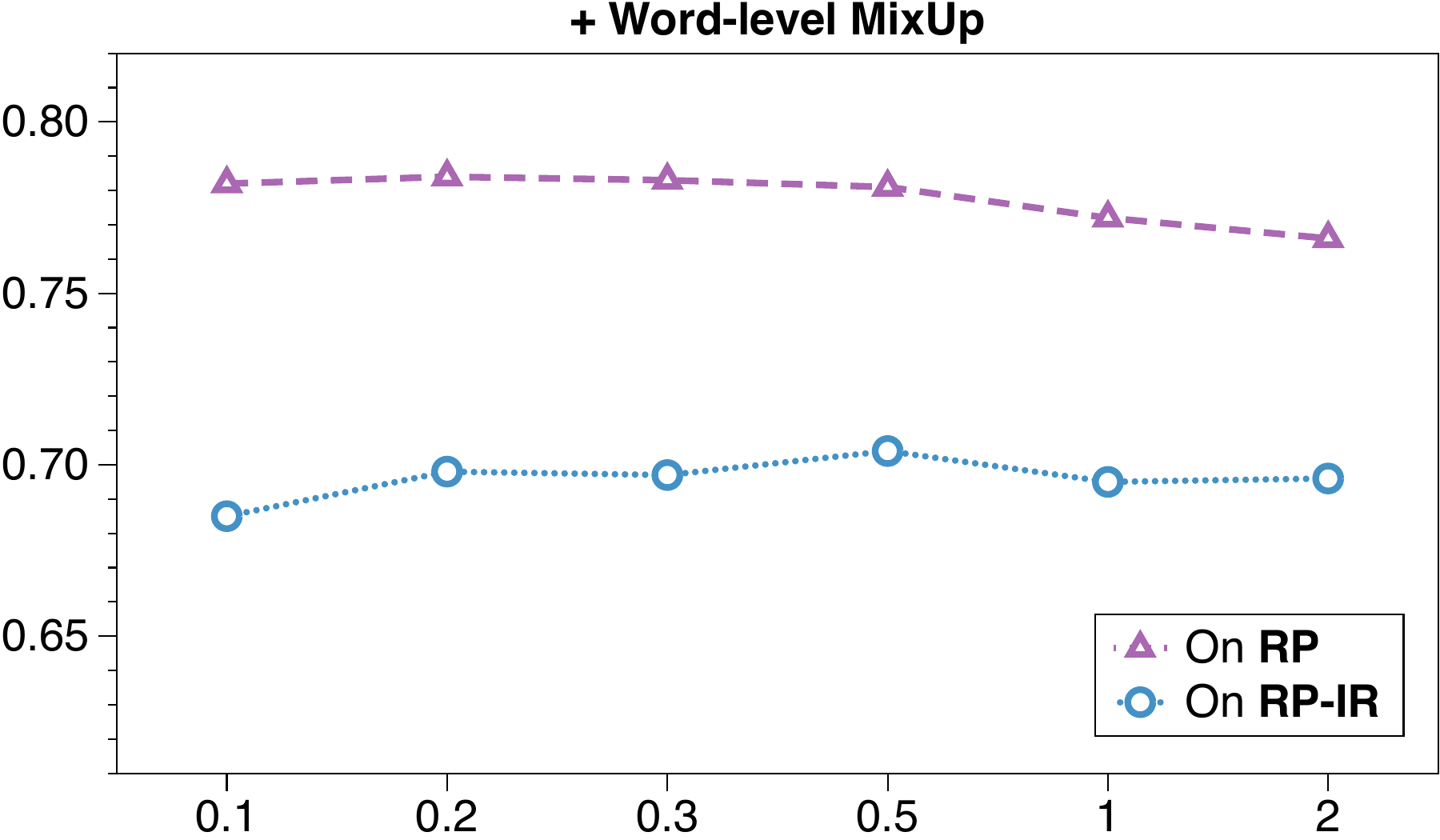}
}
\subfigure{
\includegraphics[width=7.3cm]{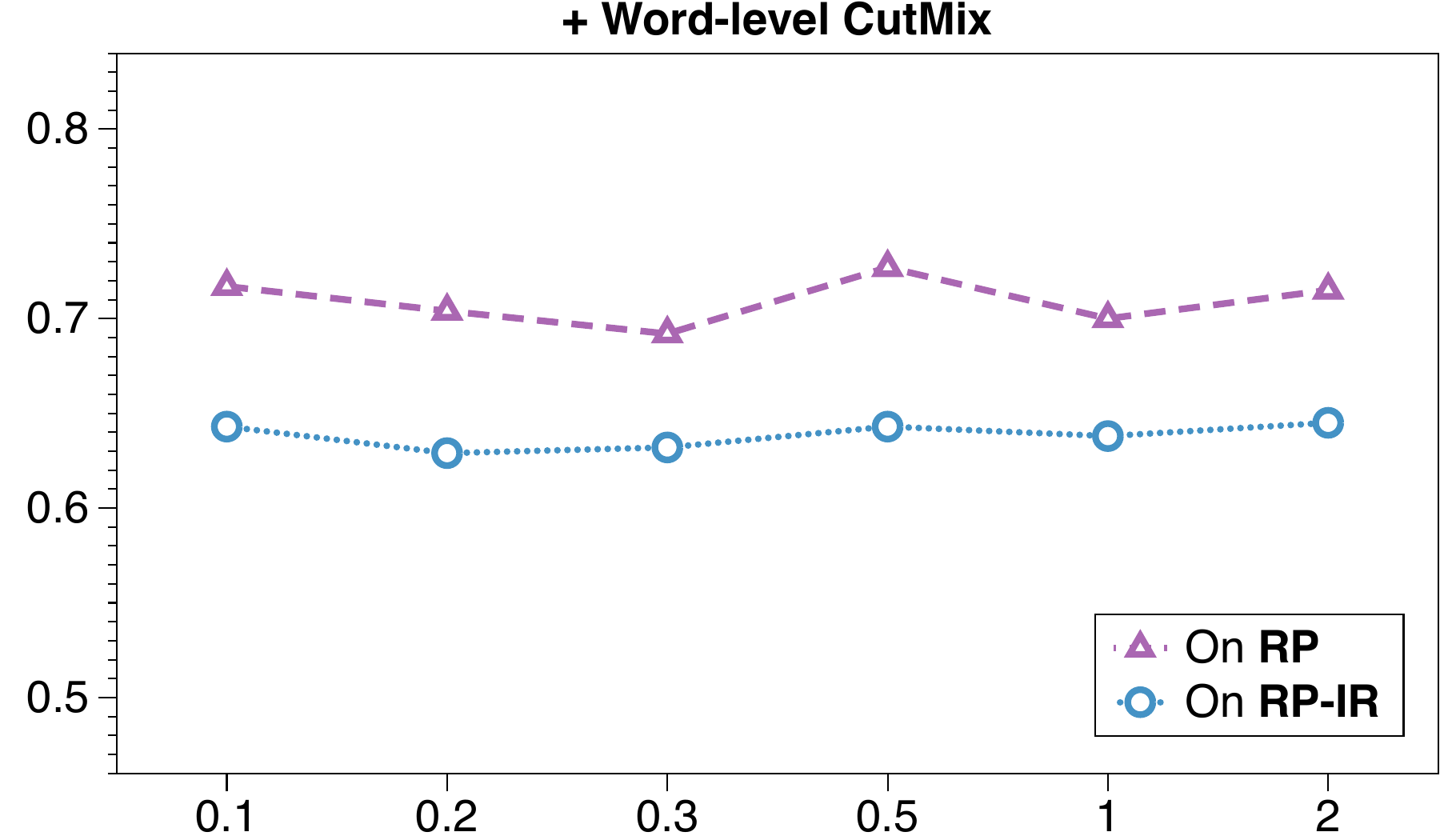}
}\\
\subfigure{
\includegraphics[width=7.3cm]{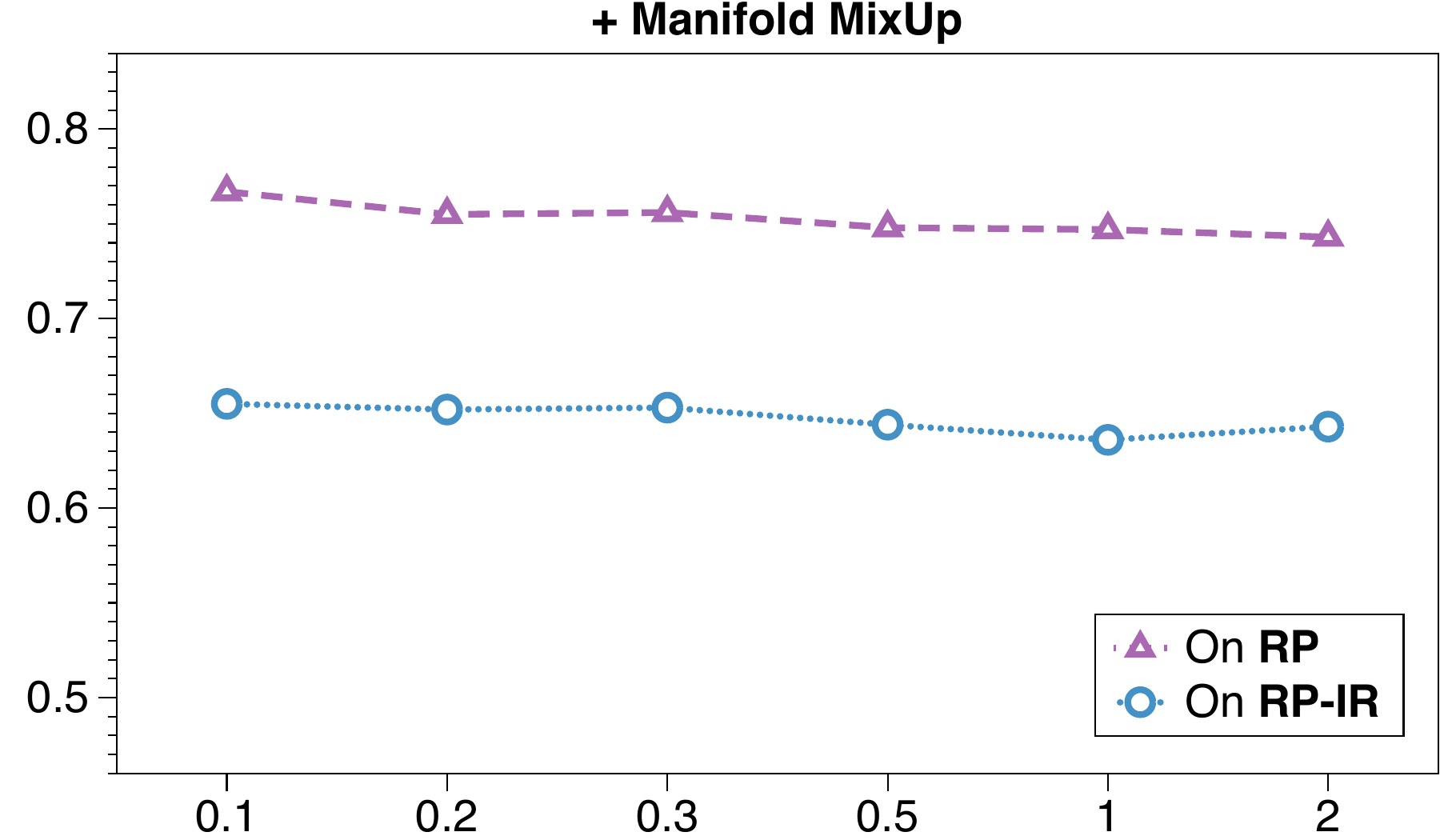}
}
\subfigure{
\includegraphics[width=7.3cm]{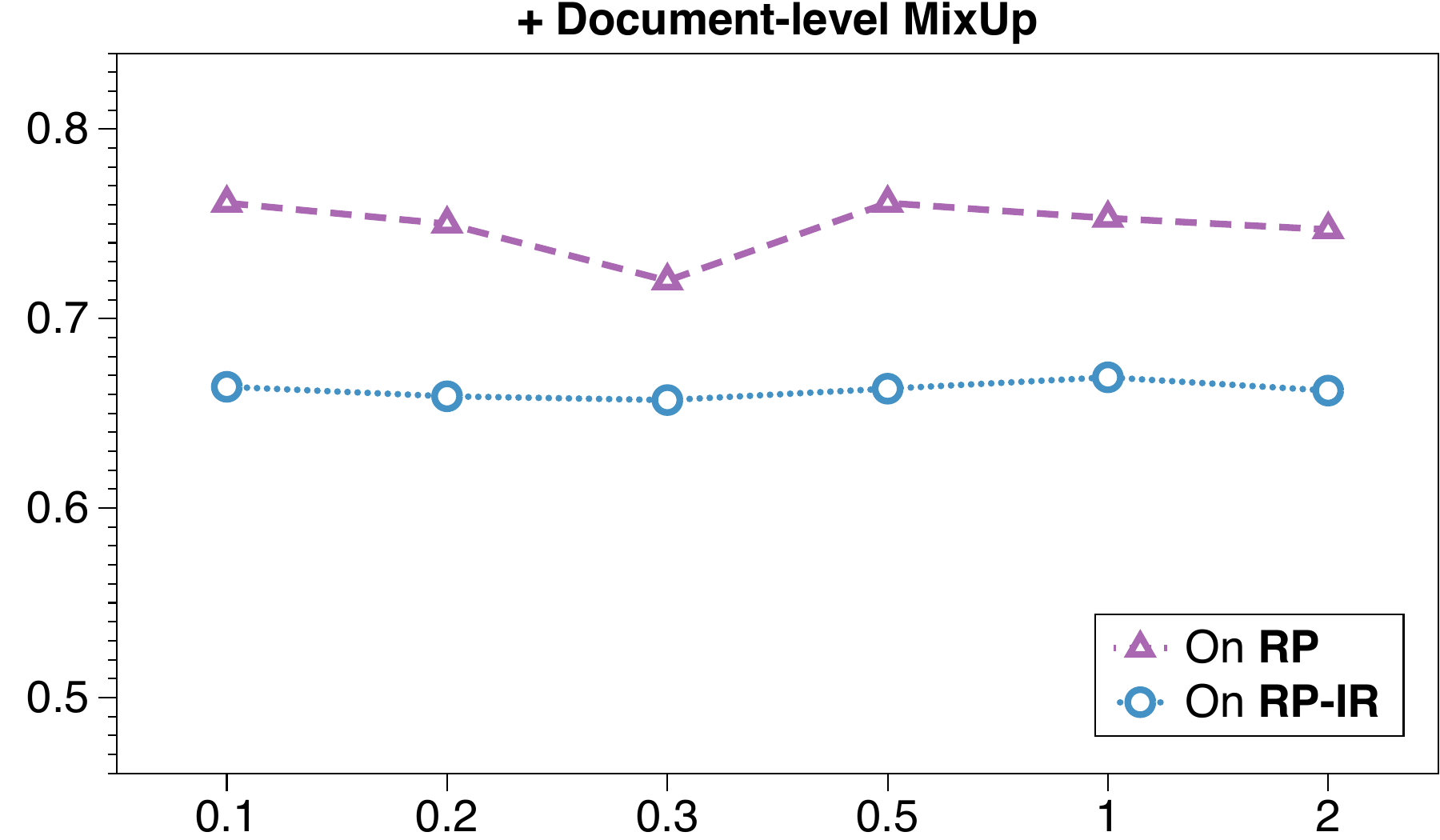}
}
\caption{Impact of different $\alpha$ selection in the Beta distribution  in terms of F1 Score.}
\label{figure.hyper}
\end{figure*}


\subsection{Hyperparameter Study}
We conduct a series of experiments with the primary aim of evaluating the impact of the critical hyperparameters, denoted as $\alpha$, of the Beta distribution depicted in Equations~\ref{mix}. The outcomes of these experiments are comprehensively reported in Figure~\ref{figure.hyper}.
Upon examination of the figure, it becomes evident that the hyperparameter $\alpha$ exercises a modest influence on the performance of the model. 
For comparative analysis, we employed the optimal $\alpha$ setting, thereby ensuring that the comparison is predicated on the most effective parameter configuration.


%% file: conclusion.tex
In this study, we introduced \model, a \fullmodel\ tailored specifically for topic inference tasks applied to the real-world imbalanced research proposal dataset. The core concept behind \model\ hinges upon three essential elements: 1) individualized modeling of each textual component within a research proposal, 2) adoption of a hierarchical multilabel classification approach for topic path inference, and 3) selective blending of input semantic features and prior prediction results during the training phase to generate high-quality pseudo research proposals, thereby addressing the interdisciplinary-non-interdisciplinary imbalance.
Our ablation study provided empirical evidence attesting to the practicality and utility of each component of \model. In addition, we evaluated four distinct selective interpolation strategies. The analysis revealed that each strategy conferred a marked improvement in model performance. We identified the most efficacious strategy through comprehensive and level-wise comparison - the Word-level MixUp.
Further exploration of the mechanisms underpinning our selective interpolation strategy allowed us to understand its impact on model performance, as observed through variations in training loss, test loss, and the attention mechanism of input text. This strategy effectively mitigates the data imbalance issue and enhances model generalization during training, providing a consistent training loss.
Moreover, applying the MixUp method at the word level yields balanced attention to domain-related keywords. This finding aligns with the performance comparison results.
Collectively, the performance enhancements brought about by \model\ showcase its potential for real-world applications and signify a promising avenue for improving peer-reviewer assignment processes.

%% file: related.tex


\subsection{Imbalance Learning}
The class imbalanced problem is common in real-world scenarios and has become a popular topic of research~\cite{johnson2019survey}. 
The mainstream imbalance learning algorithms can be divided into three categories: data-level, algorithm-level, and hybrid methods. 
Usually, the data-level methods adjust class sizes by down-sampling, over-sampling~\cite{mullick2019generative,kim2019imbalanced}, or generate pseudo data via generative model~\cite{cai2023resolving}.
The algorithm-level methods aim to directly increase the importance of minority classes through appropriate penalty functions~\cite{ling2008cost,puthiya2014optimizing}. 
Lastly, hybrid systems strategically combine sampling with algorithmic methods~\cite{krawczyk2016learning}.
However, there is little research on deep learning with class imbalanced data~\cite{anand1993improved}.
In this paper, we mainly focus on solving the hierarchical multi-label imbalanced problem for the research proposal data with Mixup-based algorithms ~\cite{Zhang2018mix}.

\subsection{Mixup}
    Mixup-based methods have proven superior for improving the generalization and robustness of deep neural networks~\cite{zhang2020does} by interpolating features and labels between two random samples. 
    Mixup~\cite{Zhang2018mix} and its numerous variants (e.g. Manifold Mixup~\cite{verma2019manifold}, Cutmix~\cite{Yun2019}, etc.), as data augmentation methods, have not only achieved notable success in a wide range of machine learning problems such as supervised learning~\cite{Zhang2018mix}, semi-supervised learning~\cite{berthelot2019mixmatch,verma2022interpolation}, adversarial learning~\cite{archambault2019mixup}, but also adapted to different data forms such as images~\cite{chen2021transmix}, texts~\cite{Xia2018,Yoon2021}, graphs~\cite{verma2021graphmix}, and speech~\cite{zhu2019mixup}. 
    Notably, to alleviate the problem of class imbalance in the dataset, a series of methods~\cite{galdran2021balanced,kabra2020mixboost,chou2020remix,cheng2020advaug,kwon2023explainability} employ Mixup to augment the data. 
    Despite this, there has not been any research on using MixUp to solve the class imbalance problem in hierarchical multi-label classification

%% file: main.bbl

\begin{thebibliography}{56}


\ifx \showCODEN    \undefined \def \showCODEN     #1{\unskip}     \fi
\ifx \showDOI      \undefined \def \showDOI       #1{#1}\fi
\ifx \showISBNx    \undefined \def \showISBNx     #1{\unskip}     \fi
\ifx \showISBNxiii \undefined \def \showISBNxiii  #1{\unskip}     \fi
\ifx \showISSN     \undefined \def \showISSN      #1{\unskip}     \fi
\ifx \showLCCN     \undefined \def \showLCCN      #1{\unskip}     \fi
\ifx \shownote     \undefined \def \shownote      #1{#1}          \fi
\ifx \showarticletitle \undefined \def \showarticletitle #1{#1}   \fi
\ifx \showURL      \undefined \def \showURL       {\relax}        \fi
\providecommand\bibfield[2]{#2}
\providecommand\bibinfo[2]{#2}
\providecommand\natexlab[1]{#1}
\providecommand\showeprint[2][]{arXiv:#2}

\bibitem[Anand et~al\mbox{.}(1993)]%
        {anand1993improved}
\bibfield{author}{\bibinfo{person}{Rangachari Anand}, \bibinfo{person}{Kishan~G Mehrotra}, \bibinfo{person}{Chilukuri~K Mohan}, {and} \bibinfo{person}{Sanjay Ranka}.} \bibinfo{year}{1993}\natexlab{}.
\newblock \showarticletitle{An improved algorithm for neural network classification of imbalanced training sets}.
\newblock \bibinfo{journal}{\emph{IEEE Transactions on Neural Networks}} \bibinfo{volume}{4}, \bibinfo{number}{6} (\bibinfo{year}{1993}), \bibinfo{pages}{962--969}.
\newblock


\bibitem[Archambault et~al\mbox{.}(2019)]%
        {archambault2019mixup}
\bibfield{author}{\bibinfo{person}{Guillaume~P Archambault}, \bibinfo{person}{Yongyi Mao}, \bibinfo{person}{Hongyu Guo}, {and} \bibinfo{person}{Richong Zhang}.} \bibinfo{year}{2019}\natexlab{}.
\newblock \showarticletitle{Mixup as directional adversarial training}.
\newblock \bibinfo{journal}{\emph{arXiv preprint arXiv:1906.06875}} (\bibinfo{year}{2019}).
\newblock


\bibitem[Berthelot et~al\mbox{.}(2019)]%
        {berthelot2019mixmatch}
\bibfield{author}{\bibinfo{person}{David Berthelot}, \bibinfo{person}{Nicholas Carlini}, \bibinfo{person}{Ian Goodfellow}, \bibinfo{person}{Nicolas Papernot}, \bibinfo{person}{Avital Oliver}, {and} \bibinfo{person}{Colin~A Raffel}.} \bibinfo{year}{2019}\natexlab{}.
\newblock \showarticletitle{Mixmatch: A holistic approach to semi-supervised learning}.
\newblock \bibinfo{journal}{\emph{Advances in Neural Information Processing Systems}}  \bibinfo{volume}{32} (\bibinfo{year}{2019}).
\newblock


\bibitem[Bojanowski et~al\mbox{.}(2017)]%
        {textrcnn}
\bibfield{author}{\bibinfo{person}{Piotr Bojanowski}, \bibinfo{person}{Edouard Grave}, \bibinfo{person}{Armand Joulin}, {and} \bibinfo{person}{Tomas Mikolov}.} \bibinfo{year}{2017}\natexlab{}.
\newblock \showarticletitle{{Enriching Word Vectors with Subword Information}}.
\newblock \bibinfo{journal}{\emph{Transactions of the Association for Computational Linguistics}}  \bibinfo{volume}{5} (\bibinfo{year}{2017}), \bibinfo{pages}{135--146}.
\newblock
\showISSN{2307-387X}
\urldef\tempurl%
\url{https://doi.org/10.1162/tacl_a_00051}
\showDOI{\tempurl}
\showeprint[arxiv]{1607.04606}


\bibitem[Cai et~al\mbox{.}(2023)]%
        {cai2023resolving}
\bibfield{author}{\bibinfo{person}{Xunxin Cai}, \bibinfo{person}{Meng Xiao}, \bibinfo{person}{Zhiyuan Ning}, {and} \bibinfo{person}{Yuanchun Zhou}.} \bibinfo{year}{2023}\natexlab{}.
\newblock \showarticletitle{Resolving the Imbalance Issue in Hierarchical Disciplinary Topic Inference via LLM-based Data Augmentation}.
\newblock \bibinfo{journal}{\emph{2023 IEEE International Conference on Data Mining Workshops (ICDMW)}} (\bibinfo{year}{2023}).
\newblock


\bibitem[Chen et~al\mbox{.}(2021)]%
        {chen2021transmix}
\bibfield{author}{\bibinfo{person}{Jie-Neng Chen}, \bibinfo{person}{Shuyang Sun}, \bibinfo{person}{Ju He}, \bibinfo{person}{Philip Torr}, \bibinfo{person}{Alan Yuille}, {and} \bibinfo{person}{Song Bai}.} \bibinfo{year}{2021}\natexlab{}.
\newblock \showarticletitle{TransMix: Attend to Mix for Vision Transformers}.
\newblock \bibinfo{journal}{\emph{arXiv preprint arXiv:2111.09833}} (\bibinfo{year}{2021}).
\newblock


\bibitem[Chen et~al\mbox{.}(2023)]%
        {10.1145/3543507.3583355}
\bibfield{author}{\bibinfo{person}{Xiao Chen}, \bibinfo{person}{Wenqi Fan}, \bibinfo{person}{Jingfan Chen}, \bibinfo{person}{Haochen Liu}, \bibinfo{person}{Zitao Liu}, \bibinfo{person}{Zhaoxiang Zhang}, {and} \bibinfo{person}{Qing Li}.} \bibinfo{year}{2023}\natexlab{}.
\newblock \showarticletitle{Fairly Adaptive Negative Sampling for Recommendations}. In \bibinfo{booktitle}{\emph{Proceedings of the ACM Web Conference 2023}} (Austin, TX, USA) \emph{(\bibinfo{series}{WWW '23})}. \bibinfo{publisher}{Association for Computing Machinery}, \bibinfo{address}{New York, NY, USA}, \bibinfo{pages}{3723–3733}.
\newblock
\showISBNx{9781450394161}
\urldef\tempurl%
\url{https://doi.org/10.1145/3543507.3583355}
\showDOI{\tempurl}


\bibitem[Cheng et~al\mbox{.}(2020)]%
        {cheng2020advaug}
\bibfield{author}{\bibinfo{person}{Yong Cheng}, \bibinfo{person}{Lu Jiang}, \bibinfo{person}{Wolfgang Macherey}, {and} \bibinfo{person}{Jacob Eisenstein}.} \bibinfo{year}{2020}\natexlab{}.
\newblock \showarticletitle{Advaug: Robust adversarial augmentation for neural machine translation}.
\newblock \bibinfo{journal}{\emph{arXiv preprint arXiv:2006.11834}} (\bibinfo{year}{2020}).
\newblock


\bibitem[Chiang et~al\mbox{.}(2023)]%
        {vicuna2023}
\bibfield{author}{\bibinfo{person}{Wei-Lin Chiang}, \bibinfo{person}{Zhuohan Li}, \bibinfo{person}{Zi Lin}, \bibinfo{person}{Ying Sheng}, \bibinfo{person}{Zhanghao Wu}, \bibinfo{person}{Hao Zhang}, \bibinfo{person}{Lianmin Zheng}, \bibinfo{person}{Siyuan Zhuang}, \bibinfo{person}{Yonghao Zhuang}, \bibinfo{person}{Joseph~E. Gonzalez}, \bibinfo{person}{Ion Stoica}, {and} \bibinfo{person}{Eric~P. Xing}.} \bibinfo{year}{2023}\natexlab{}.
\newblock \bibinfo{title}{Vicuna: An Open-Source Chatbot Impressing GPT-4 with 90\%* ChatGPT Quality}.
\newblock
\newblock
\urldef\tempurl%
\url{https://lmsys.org/blog/2023-03-30-vicuna/}
\showURL{%
\tempurl}


\bibitem[Chou et~al\mbox{.}(2020)]%
        {chou2020remix}
\bibfield{author}{\bibinfo{person}{Hsin-Ping Chou}, \bibinfo{person}{Shih-Chieh Chang}, \bibinfo{person}{Jia-Yu Pan}, \bibinfo{person}{Wei Wei}, {and} \bibinfo{person}{Da-Cheng Juan}.} \bibinfo{year}{2020}\natexlab{}.
\newblock \showarticletitle{Remix: rebalanced mixup}. In \bibinfo{booktitle}{\emph{European Conference on Computer Vision}}. Springer, \bibinfo{pages}{95--110}.
\newblock


\bibitem[Dosovitskiy et~al\mbox{.}(2020)]%
        {dosovitskiy2020image}
\bibfield{author}{\bibinfo{person}{Alexey Dosovitskiy}, \bibinfo{person}{Lucas Beyer}, \bibinfo{person}{Alexander Kolesnikov}, \bibinfo{person}{Dirk Weissenborn}, \bibinfo{person}{Xiaohua Zhai}, \bibinfo{person}{Thomas Unterthiner}, \bibinfo{person}{Mostafa Dehghani}, \bibinfo{person}{Matthias Minderer}, \bibinfo{person}{Georg Heigold}, \bibinfo{person}{Sylvain Gelly}, {et~al\mbox{.}}} \bibinfo{year}{2020}\natexlab{}.
\newblock \showarticletitle{An image is worth 16x16 words: Transformers for image recognition at scale}.
\newblock \bibinfo{journal}{\emph{arXiv preprint arXiv:2010.11929}} (\bibinfo{year}{2020}).
\newblock


\bibitem[Galdran et~al\mbox{.}(2021)]%
        {galdran2021balanced}
\bibfield{author}{\bibinfo{person}{Adrian Galdran}, \bibinfo{person}{Gustavo Carneiro}, {and} \bibinfo{person}{Miguel~A Gonz{\'a}lez~Ballester}.} \bibinfo{year}{2021}\natexlab{}.
\newblock \showarticletitle{Balanced-MixUp for Highly Imbalanced Medical Image Classification}. In \bibinfo{booktitle}{\emph{International Conference on Medical Image Computing and Computer-Assisted Intervention}}. Springer, \bibinfo{pages}{323--333}.
\newblock


\bibitem[Gibaja and Ventura(2014)]%
        {metric1}
\bibfield{author}{\bibinfo{person}{Eva Gibaja} {and} \bibinfo{person}{Sebasti{\'a}n Ventura}.} \bibinfo{year}{2014}\natexlab{}.
\newblock \showarticletitle{Multi-label learning: a review of the state of the art and ongoing research}.
\newblock \bibinfo{journal}{\emph{Wiley Interdisciplinary Reviews: Data Mining and Knowledge Discovery}} \bibinfo{volume}{4}, \bibinfo{number}{6} (\bibinfo{year}{2014}), \bibinfo{pages}{411--444}.
\newblock


\bibitem[Han et~al\mbox{.}(2021)]%
        {han2021transformer}
\bibfield{author}{\bibinfo{person}{Kai Han}, \bibinfo{person}{An Xiao}, \bibinfo{person}{Enhua Wu}, \bibinfo{person}{Jianyuan Guo}, \bibinfo{person}{Chunjing Xu}, {and} \bibinfo{person}{Yunhe Wang}.} \bibinfo{year}{2021}\natexlab{}.
\newblock \showarticletitle{Transformer in transformer}.
\newblock \bibinfo{journal}{\emph{arXiv preprint arXiv:2103.00112}} (\bibinfo{year}{2021}).
\newblock


\bibitem[Huang et~al\mbox{.}(2019)]%
        {huang2019hierarchical}
\bibfield{author}{\bibinfo{person}{Wei Huang}, \bibinfo{person}{Enhong Chen}, \bibinfo{person}{Qi Liu}, \bibinfo{person}{Yuying Chen}, \bibinfo{person}{Zai Huang}, \bibinfo{person}{Yang Liu}, \bibinfo{person}{Zhou Zhao}, \bibinfo{person}{Dan Zhang}, {and} \bibinfo{person}{Shijin Wang}.} \bibinfo{year}{2019}\natexlab{}.
\newblock \showarticletitle{Hierarchical multi-label text classification: An attention-based recurrent network approach}. In \bibinfo{booktitle}{\emph{Proceedings of the 28th ACM International Conference on Information and Knowledge Management}}. \bibinfo{pages}{1051--1060}.
\newblock


\bibitem[Johnson and Khoshgoftaar(2019)]%
        {johnson2019survey}
\bibfield{author}{\bibinfo{person}{Justin~M Johnson} {and} \bibinfo{person}{Taghi~M Khoshgoftaar}.} \bibinfo{year}{2019}\natexlab{}.
\newblock \showarticletitle{Survey on deep learning with class imbalance}.
\newblock \bibinfo{journal}{\emph{Journal of Big Data}} \bibinfo{volume}{6}, \bibinfo{number}{1} (\bibinfo{year}{2019}), \bibinfo{pages}{1--54}.
\newblock


\bibitem[Johnson and Zhang(2017)]%
        {dpcnn}
\bibfield{author}{\bibinfo{person}{Rie Johnson} {and} \bibinfo{person}{Tong Zhang}.} \bibinfo{year}{2017}\natexlab{}.
\newblock \showarticletitle{{Deep pyramid convolutional neural networks for text categorization}}.
\newblock \bibinfo{journal}{\emph{ACL 2017 - 55th Annual Meeting of the Association for Computational Linguistics, Proceedings of the Conference (Long Papers)}}  \bibinfo{volume}{1} (\bibinfo{year}{2017}), \bibinfo{pages}{562--570}.
\newblock
\showISBNx{9781945626753}
\urldef\tempurl%
\url{https://doi.org/10.18653/v1/P17-1052}
\showDOI{\tempurl}


\bibitem[Joulin et~al\mbox{.}(2017)]%
        {fasttext1}
\bibfield{author}{\bibinfo{person}{Armand Joulin}, \bibinfo{person}{Edouard Grave}, \bibinfo{person}{Piotr Bojanowski}, {and} \bibinfo{person}{Tomas Mikolov}.} \bibinfo{year}{2017}\natexlab{}.
\newblock \showarticletitle{{Bag of tricks for efficient text classification}}.
\newblock \bibinfo{journal}{\emph{15th Conference of the European Chapter of the Association for Computational Linguistics, EACL 2017 - Proceedings of Conference}}  \bibinfo{volume}{2} (\bibinfo{year}{2017}), \bibinfo{pages}{427--431}.
\newblock
\showISBNx{9781510838604}
\urldef\tempurl%
\url{https://doi.org/10.18653/v1/e17-2068}
\showDOI{\tempurl}
\showeprint[arxiv]{1607.01759}


\bibitem[Kabra et~al\mbox{.}(2020)]%
        {kabra2020mixboost}
\bibfield{author}{\bibinfo{person}{Anubha Kabra}, \bibinfo{person}{Ayush Chopra}, \bibinfo{person}{Nikaash Puri}, \bibinfo{person}{Pinkesh Badjatiya}, \bibinfo{person}{Sukriti Verma}, \bibinfo{person}{Piyush Gupta}, {et~al\mbox{.}}} \bibinfo{year}{2020}\natexlab{}.
\newblock \showarticletitle{MixBoost: Synthetic Oversampling with Boosted Mixup for Handling Extreme Imbalance}.
\newblock \bibinfo{journal}{\emph{arXiv preprint arXiv:2009.01571}} (\bibinfo{year}{2020}).
\newblock


\bibitem[Kim et~al\mbox{.}(2019)]%
        {kim2019imbalanced}
\bibfield{author}{\bibinfo{person}{Jaehyung Kim}, \bibinfo{person}{Jongheon Jeong}, {and} \bibinfo{person}{Jinwoo Shin}.} \bibinfo{year}{2019}\natexlab{}.
\newblock \showarticletitle{Imbalanced classification via adversarial minority over-sampling}.
\newblock  (\bibinfo{year}{2019}).
\newblock


\bibitem[Kim(2014)]%
        {Kim2014}
\bibfield{author}{\bibinfo{person}{Yoon Kim}.} \bibinfo{year}{2014}\natexlab{}.
\newblock \showarticletitle{{Convolutional neural networks for sentence classification}}.
\newblock \bibinfo{journal}{\emph{EMNLP 2014 - 2014 Conference on Empirical Methods in Natural Language Processing, Proceedings of the Conference}} (\bibinfo{year}{2014}), \bibinfo{pages}{1746--1751}.
\newblock
\showISBNx{9781937284961}
\urldef\tempurl%
\url{https://doi.org/10.3115/v1/d14-1181}
\showDOI{\tempurl}
\showeprint[arxiv]{1408.5882}


\bibitem[Krawczyk(2016)]%
        {krawczyk2016learning}
\bibfield{author}{\bibinfo{person}{Bartosz Krawczyk}.} \bibinfo{year}{2016}\natexlab{}.
\newblock \showarticletitle{Learning from imbalanced data: open challenges and future directions}.
\newblock \bibinfo{journal}{\emph{Progress in Artificial Intelligence}} \bibinfo{volume}{5}, \bibinfo{number}{4} (\bibinfo{year}{2016}), \bibinfo{pages}{221--232}.
\newblock


\bibitem[Kwon and Lee(2023)]%
        {kwon2023explainability}
\bibfield{author}{\bibinfo{person}{Soonki Kwon} {and} \bibinfo{person}{Younghoon Lee}.} \bibinfo{year}{2023}\natexlab{}.
\newblock \showarticletitle{Explainability-based mix-up approach for text data augmentation}.
\newblock \bibinfo{journal}{\emph{ACM Transactions on Knowledge Discovery from Data}} \bibinfo{volume}{17}, \bibinfo{number}{1} (\bibinfo{year}{2023}), \bibinfo{pages}{1--14}.
\newblock


\bibitem[Ling and Sheng(2008)]%
        {ling2008cost}
\bibfield{author}{\bibinfo{person}{Charles~X Ling} {and} \bibinfo{person}{Victor~S Sheng}.} \bibinfo{year}{2008}\natexlab{}.
\newblock \showarticletitle{Cost-sensitive learning and the class imbalance problem}.
\newblock \bibinfo{journal}{\emph{Encyclopedia of machine learning}}  \bibinfo{volume}{2011} (\bibinfo{year}{2008}), \bibinfo{pages}{231--235}.
\newblock


\bibitem[Liu et~al\mbox{.}(2016)]%
        {textrnn}
\bibfield{author}{\bibinfo{person}{Pengfei Liu}, \bibinfo{person}{Xipeng Qiu}, {and} \bibinfo{person}{Huang Xuanjing}.} \bibinfo{year}{2016}\natexlab{}.
\newblock \showarticletitle{{Recurrent neural network for text classification with multi-task learning}}.
\newblock \bibinfo{journal}{\emph{IJCAI International Joint Conference on Artificial Intelligence}}  \bibinfo{volume}{2016-Janua} (\bibinfo{year}{2016}), \bibinfo{pages}{2873--2879}.
\newblock
\showISSN{10450823}
\showeprint[arxiv]{1605.05101}


\bibitem[Mao et~al\mbox{.}(2019)]%
        {mao2019hierarchical}
\bibfield{author}{\bibinfo{person}{Yuning Mao}, \bibinfo{person}{Jingjing Tian}, \bibinfo{person}{Jiawei Han}, {and} \bibinfo{person}{Xiang Ren}.} \bibinfo{year}{2019}\natexlab{}.
\newblock \showarticletitle{Hierarchical text classification with reinforced label assignment}.
\newblock \bibinfo{journal}{\emph{arXiv preprint arXiv:1908.10419}} (\bibinfo{year}{2019}).
\newblock


\bibitem[{Mikolov} et~al\mbox{.}(2013)]%
        {mikolov2013distributed}
\bibfield{author}{\bibinfo{person}{Tomas {Mikolov}}, \bibinfo{person}{Ilya {Sutskever}}, \bibinfo{person}{Kai {Chen}}, \bibinfo{person}{Greg~S {Corrado}}, {and} \bibinfo{person}{Jeff {Dean}}.} \bibinfo{year}{2013}\natexlab{}.
\newblock \showarticletitle{Distributed Representations of Words and Phrases and their Compositionality}. In \bibinfo{booktitle}{\emph{Advances in Neural Information Processing Systems 26}}, Vol.~\bibinfo{volume}{26}. \bibinfo{pages}{3111--3119}.
\newblock


\bibitem[Mullick et~al\mbox{.}(2019)]%
        {mullick2019generative}
\bibfield{author}{\bibinfo{person}{Sankha~Subhra Mullick}, \bibinfo{person}{Shounak Datta}, {and} \bibinfo{person}{Swagatam Das}.} \bibinfo{year}{2019}\natexlab{}.
\newblock \showarticletitle{Generative adversarial minority oversampling}. In \bibinfo{booktitle}{\emph{Proceedings of the IEEE/CVF International Conference on Computer Vision}}. \bibinfo{pages}{1695--1704}.
\newblock


\bibitem[Peng et~al\mbox{.}(2019)]%
        {peng2019hierarchical}
\bibfield{author}{\bibinfo{person}{Hao Peng}, \bibinfo{person}{Jianxin Li}, \bibinfo{person}{Senzhang Wang}, \bibinfo{person}{Lihong Wang}, \bibinfo{person}{Qiran Gong}, \bibinfo{person}{Renyu Yang}, \bibinfo{person}{Bo Li}, \bibinfo{person}{Philip Yu}, {and} \bibinfo{person}{Lifang He}.} \bibinfo{year}{2019}\natexlab{}.
\newblock \showarticletitle{Hierarchical taxonomy-aware and attentional graph capsule RCNNs for large-scale multi-label text classification}.
\newblock \bibinfo{journal}{\emph{IEEE Transactions on Knowledge and Data Engineering}} (\bibinfo{year}{2019}).
\newblock


\bibitem[Puthiya~Parambath et~al\mbox{.}(2014)]%
        {puthiya2014optimizing}
\bibfield{author}{\bibinfo{person}{Shameem Puthiya~Parambath}, \bibinfo{person}{Nicolas Usunier}, {and} \bibinfo{person}{Yves Grandvalet}.} \bibinfo{year}{2014}\natexlab{}.
\newblock \showarticletitle{Optimizing F-measures by cost-sensitive classification}.
\newblock \bibinfo{journal}{\emph{Advances in neural information processing systems}}  \bibinfo{volume}{27} (\bibinfo{year}{2014}).
\newblock


\bibitem[Qiao et~al\mbox{.}(2022)]%
        {qiao2022rpt}
\bibfield{author}{\bibinfo{person}{Ziyue Qiao}, \bibinfo{person}{Yanjie Fu}, \bibinfo{person}{Pengyang Wang}, \bibinfo{person}{Meng Xiao}, \bibinfo{person}{Zhiyuan Ning}, \bibinfo{person}{Denghui Zhang}, \bibinfo{person}{Yi Du}, {and} \bibinfo{person}{Yuanchun Zhou}.} \bibinfo{year}{2022}\natexlab{}.
\newblock \showarticletitle{Rpt: toward transferable model on heterogeneous researcher data via pre-training}.
\newblock \bibinfo{journal}{\emph{IEEE Transactions on Big Data}} \bibinfo{volume}{9}, \bibinfo{number}{1} (\bibinfo{year}{2022}), \bibinfo{pages}{186--199}.
\newblock


\bibitem[Qiao et~al\mbox{.}(2023)]%
        {qiao2023semi}
\bibfield{author}{\bibinfo{person}{Ziyue Qiao}, \bibinfo{person}{Xiao Luo}, \bibinfo{person}{Meng Xiao}, \bibinfo{person}{Hao Dong}, \bibinfo{person}{Yuanchun Zhou}, {and} \bibinfo{person}{Hui Xiong}.} \bibinfo{year}{2023}\natexlab{}.
\newblock \showarticletitle{Semi-supervised domain adaptation in graph transfer learning}. In \bibinfo{booktitle}{\emph{Proceedings of the Thirty-Second International Joint Conference on Artificial Intelligence}}. \bibinfo{pages}{2279--2287}.
\newblock


\bibitem[Touvron et~al\mbox{.}(2023)]%
        {touvron2023llama}
\bibfield{author}{\bibinfo{person}{Hugo Touvron}, \bibinfo{person}{Thibaut Lavril}, \bibinfo{person}{Gautier Izacard}, \bibinfo{person}{Xavier Martinet}, \bibinfo{person}{Marie-Anne Lachaux}, \bibinfo{person}{Timoth{\'e}e Lacroix}, \bibinfo{person}{Baptiste Rozi{\`e}re}, \bibinfo{person}{Naman Goyal}, \bibinfo{person}{Eric Hambro}, \bibinfo{person}{Faisal Azhar}, {et~al\mbox{.}}} \bibinfo{year}{2023}\natexlab{}.
\newblock \showarticletitle{Llama: Open and efficient foundation language models}.
\newblock \bibinfo{journal}{\emph{arXiv preprint arXiv:2302.13971}} (\bibinfo{year}{2023}).
\newblock


\bibitem[{Vaswani} et~al\mbox{.}(2017)]%
        {vaswani2017attention}
\bibfield{author}{\bibinfo{person}{Ashish {Vaswani}}, \bibinfo{person}{Noam {Shazeer}}, \bibinfo{person}{Niki {Parmar}}, \bibinfo{person}{Jakob {Uszkoreit}}, \bibinfo{person}{Llion {Jones}}, \bibinfo{person}{Aidan~N. {Gomez}}, \bibinfo{person}{Lukasz {Kaiser}}, {and} \bibinfo{person}{Illia {Polosukhin}}.} \bibinfo{year}{2017}\natexlab{}.
\newblock \showarticletitle{Attention is All You Need}. In \bibinfo{booktitle}{\emph{Proceedings of the 31st International Conference on Neural Information Processing Systems}}, Vol.~\bibinfo{volume}{30}. \bibinfo{pages}{5998--6008}.
\newblock


\bibitem[Vens et~al\mbox{.}(2008)]%
        {metric2}
\bibfield{author}{\bibinfo{person}{Celine Vens}, \bibinfo{person}{Jan Struyf}, \bibinfo{person}{Leander Schietgat}, \bibinfo{person}{Sa{\v{s}}o D{\v{z}}eroski}, {and} \bibinfo{person}{Hendrik Blockeel}.} \bibinfo{year}{2008}\natexlab{}.
\newblock \showarticletitle{Decision trees for hierarchical multi-label classification}.
\newblock \bibinfo{journal}{\emph{Machine learning}} \bibinfo{volume}{73}, \bibinfo{number}{2} (\bibinfo{year}{2008}), \bibinfo{pages}{185}.
\newblock


\bibitem[Verma et~al\mbox{.}(2022)]%
        {verma2022interpolation}
\bibfield{author}{\bibinfo{person}{Vikas Verma}, \bibinfo{person}{Kenji Kawaguchi}, \bibinfo{person}{Alex Lamb}, \bibinfo{person}{Juho Kannala}, \bibinfo{person}{Arno Solin}, \bibinfo{person}{Yoshua Bengio}, {and} \bibinfo{person}{David Lopez-Paz}.} \bibinfo{year}{2022}\natexlab{}.
\newblock \showarticletitle{Interpolation consistency training for semi-supervised learning}.
\newblock \bibinfo{journal}{\emph{Neural Networks}}  \bibinfo{volume}{145} (\bibinfo{year}{2022}), \bibinfo{pages}{90--106}.
\newblock


\bibitem[Verma et~al\mbox{.}(2019)]%
        {verma2019manifold}
\bibfield{author}{\bibinfo{person}{Vikas Verma}, \bibinfo{person}{Alex Lamb}, \bibinfo{person}{Christopher Beckham}, \bibinfo{person}{Amir Najafi}, \bibinfo{person}{Ioannis Mitliagkas}, \bibinfo{person}{David Lopez-Paz}, {and} \bibinfo{person}{Yoshua Bengio}.} \bibinfo{year}{2019}\natexlab{}.
\newblock \showarticletitle{Manifold mixup: Better representations by interpolating hidden states}. In \bibinfo{booktitle}{\emph{International Conference on Machine Learning}}. PMLR, \bibinfo{pages}{6438--6447}.
\newblock


\bibitem[Verma et~al\mbox{.}(2021)]%
        {verma2021graphmix}
\bibfield{author}{\bibinfo{person}{Vikas Verma}, \bibinfo{person}{Meng Qu}, \bibinfo{person}{Kenji Kawaguchi}, \bibinfo{person}{Alex Lamb}, \bibinfo{person}{Yoshua Bengio}, \bibinfo{person}{Juho Kannala}, {and} \bibinfo{person}{Jian Tang}.} \bibinfo{year}{2021}\natexlab{}.
\newblock \showarticletitle{GraphMix: Improved Training of GNNs for Semi-Supervised Learning}. In \bibinfo{booktitle}{\emph{Proceedings of the AAAI Conference on Artificial Intelligence}}.
\newblock


\bibitem[Wang et~al\mbox{.}(2024)]%
        {wang2024reinforcement}
\bibfield{author}{\bibinfo{person}{Dongjie Wang}, \bibinfo{person}{Meng Xiao}, \bibinfo{person}{Min Wu}, \bibinfo{person}{Yuanchun Zhou}, \bibinfo{person}{Yanjie Fu}, {et~al\mbox{.}}} \bibinfo{year}{2024}\natexlab{}.
\newblock \showarticletitle{Reinforcement-enhanced autoregressive feature transformation: Gradient-steered search in continuous space for postfix expressions}.
\newblock \bibinfo{journal}{\emph{Advances in Neural Information Processing Systems}}  \bibinfo{volume}{36} (\bibinfo{year}{2024}).
\newblock


\bibitem[Wehrmann et~al\mbox{.}(2018)]%
        {wehrmann2018hierarchical}
\bibfield{author}{\bibinfo{person}{Jonatas Wehrmann}, \bibinfo{person}{Ricardo Cerri}, {and} \bibinfo{person}{Rodrigo Barros}.} \bibinfo{year}{2018}\natexlab{}.
\newblock \showarticletitle{Hierarchical multi-label classification networks}. In \bibinfo{booktitle}{\emph{International Conference on Machine Learning}}. PMLR, \bibinfo{pages}{5075--5084}.
\newblock


\bibitem[Wu et~al\mbox{.}(2022)]%
        {wu2022label}
\bibfield{author}{\bibinfo{person}{Changxing Wu}, \bibinfo{person}{Liuwen Cao}, \bibinfo{person}{Yubin Ge}, \bibinfo{person}{Yang Liu}, \bibinfo{person}{Min Zhang}, {and} \bibinfo{person}{Jinsong Su}.} \bibinfo{year}{2022}\natexlab{}.
\newblock \showarticletitle{A Label Dependence-aware Sequence Generation Model for Multi-level Implicit Discourse Relation Recognition}. In \bibinfo{booktitle}{\emph{Proceedings of the AAAI Conference on Artificial Intelligence}}, Vol.~\bibinfo{volume}{36}. \bibinfo{pages}{11486--11494}.
\newblock


\bibitem[Xia(2018)]%
        {Xia2018}
\bibfield{author}{\bibinfo{person}{Congying Xia}.} \bibinfo{year}{2018}\natexlab{}.
\newblock \showarticletitle{{Mixup-Transformer: Dynamic Data Augmentation for NLP Tasks}}.
\newblock  (\bibinfo{year}{2018}).
\newblock
\showeprint[arxiv]{arXiv:2010.02394v2}


\bibitem[Xiao et~al\mbox{.}(2023a)]%
        {xiao2023hierarchical}
\bibfield{author}{\bibinfo{person}{Meng Xiao}, \bibinfo{person}{Ziyue Qiao}, \bibinfo{person}{Yanjie Fu}, \bibinfo{person}{Hao Dong}, \bibinfo{person}{Yi Du}, \bibinfo{person}{Pengyang Wang}, \bibinfo{person}{Hui Xiong}, {and} \bibinfo{person}{Yuanchun Zhou}.} \bibinfo{year}{2023}\natexlab{a}.
\newblock \showarticletitle{Hierarchical interdisciplinary topic detection model for research proposal classification}.
\newblock \bibinfo{journal}{\emph{IEEE Transactions on Knowledge and Data Engineering}} (\bibinfo{year}{2023}).
\newblock


\bibitem[Xiao et~al\mbox{.}(2021)]%
        {xiao2021expert}
\bibfield{author}{\bibinfo{person}{Meng Xiao}, \bibinfo{person}{Ziyue Qiao}, \bibinfo{person}{Yanjie Fu}, \bibinfo{person}{Yi Du}, {and} \bibinfo{person}{Pengyang Wang}.} \bibinfo{year}{2021}\natexlab{}.
\newblock \showarticletitle{Expert Knowledge-Guided Length-Variant Hierarchical Label Generation for Proposal Classification}.
\newblock \bibinfo{journal}{\emph{2021 IEEE International Conference on Data Mining}} (\bibinfo{year}{2021}), \bibinfo{pages}{757--766}.
\newblock


\bibitem[Xiao et~al\mbox{.}(2024)]%
        {xiao2024traceable}
\bibfield{author}{\bibinfo{person}{Meng Xiao}, \bibinfo{person}{Dongjie Wang}, \bibinfo{person}{Min Wu}, \bibinfo{person}{Kunpeng Liu}, \bibinfo{person}{Hui Xiong}, \bibinfo{person}{Yuanchun Zhou}, {and} \bibinfo{person}{Yanjie Fu}.} \bibinfo{year}{2024}\natexlab{}.
\newblock \showarticletitle{Traceable group-wise self-optimizing feature transformation learning: A dual optimization perspective}.
\newblock \bibinfo{journal}{\emph{ACM Transactions on Knowledge Discovery from Data}} \bibinfo{volume}{18}, \bibinfo{number}{4} (\bibinfo{year}{2024}), \bibinfo{pages}{1--22}.
\newblock


\bibitem[Xiao et~al\mbox{.}(2023b)]%
        {xiao2023traceable}
\bibfield{author}{\bibinfo{person}{Meng Xiao}, \bibinfo{person}{Dongjie Wang}, \bibinfo{person}{Min Wu}, \bibinfo{person}{Ziyue Qiao}, \bibinfo{person}{Pengfei Wang}, \bibinfo{person}{Kunpeng Liu}, \bibinfo{person}{Yuanchun Zhou}, {and} \bibinfo{person}{Yanjie Fu}.} \bibinfo{year}{2023}\natexlab{b}.
\newblock \showarticletitle{Traceable automatic feature transformation via cascading actor-critic agents}. In \bibinfo{booktitle}{\emph{Proceedings of the 2023 SIAM International Conference on Data Mining (SDM)}}. SIAM, \bibinfo{pages}{775--783}.
\newblock


\bibitem[Xiao et~al\mbox{.}(2023c)]%
        {xiao2023beyond}
\bibfield{author}{\bibinfo{person}{Meng Xiao}, \bibinfo{person}{Dongjie Wang}, \bibinfo{person}{Min Wu}, \bibinfo{person}{Pengfei Wang}, \bibinfo{person}{Yuanchun Zhou}, {and} \bibinfo{person}{Yanjie Fu}.} \bibinfo{year}{2023}\natexlab{c}.
\newblock \showarticletitle{Beyond discrete selection: Continuous embedding space optimization for generative feature selection}. In \bibinfo{booktitle}{\emph{2023 IEEE International Conference on Data Mining (ICDM)}}. IEEE, \bibinfo{pages}{688--697}.
\newblock


\bibitem[Ye et~al\mbox{.}(2023)]%
        {ye2023needed}
\bibfield{author}{\bibinfo{person}{Xu Ye}, \bibinfo{person}{Meng Xiao}, \bibinfo{person}{Zhiyuan Ning}, \bibinfo{person}{Weiwei Dai}, \bibinfo{person}{Wenjuan Cui}, \bibinfo{person}{Yi Du}, {and} \bibinfo{person}{Yuanchun Zhou}.} \bibinfo{year}{2023}\natexlab{}.
\newblock \showarticletitle{Needed: Introducing hierarchical transformer to eye diseases diagnosis}. In \bibinfo{booktitle}{\emph{Proceedings of the 2023 SIAM International Conference on Data Mining (SDM)}}. SIAM, \bibinfo{pages}{667--675}.
\newblock


\bibitem[Yoon et~al\mbox{.}(2021)]%
        {Yoon2021}
\bibfield{author}{\bibinfo{person}{Soyoung Yoon}, \bibinfo{person}{Gyuwan Kim}, {and} \bibinfo{person}{Kyumin Park}.} \bibinfo{year}{2021}\natexlab{}.
\newblock \showarticletitle{{SSMix: Saliency-Based Span Mixup for Text Classification}}.
\newblock  \bibinfo{number}{Fig 1} (\bibinfo{year}{2021}), \bibinfo{pages}{3225--3234}.
\newblock
\showISBNx{9781954085541}
\urldef\tempurl%
\url{https://doi.org/10.18653/v1/2021.findings-acl.285}
\showDOI{\tempurl}
\showeprint[arxiv]{2106.08062}


\bibitem[Yun et~al\mbox{.}(2019)]%
        {Yun2019}
\bibfield{author}{\bibinfo{person}{Sangdoo Yun}, \bibinfo{person}{Dongyoon Han}, \bibinfo{person}{Sanghyuk Chun}, \bibinfo{person}{Seong~Joon Oh}, \bibinfo{person}{Junsuk Choe}, {and} \bibinfo{person}{Youngjoon Yoo}.} \bibinfo{year}{2019}\natexlab{}.
\newblock \showarticletitle{{CutMix: Regularization strategy to train strong classifiers with localizable features}}.
\newblock \bibinfo{journal}{\emph{Proceedings of the IEEE International Conference on Computer Vision}} \bibinfo{volume}{2019-Octob}, \bibinfo{number}{Iccv} (\bibinfo{year}{2019}), \bibinfo{pages}{6022--6031}.
\newblock
\showISBNx{9781728148038}
\showISSN{15505499}
\urldef\tempurl%
\url{https://doi.org/10.1109/ICCV.2019.00612}
\showDOI{\tempurl}
\showeprint[arxiv]{1905.04899}


\bibitem[Zhang et~al\mbox{.}(2018)]%
        {Zhang2018mix}
\bibfield{author}{\bibinfo{person}{Hongyi Zhang}, \bibinfo{person}{Moustapha Cisse}, \bibinfo{person}{Yann~N. Dauphin}, {and} \bibinfo{person}{David Lopez-Paz}.} \bibinfo{year}{2018}\natexlab{}.
\newblock \showarticletitle{{MixUp: Beyond empirical risk minimization}}.
\newblock \bibinfo{journal}{\emph{6th International Conference on Learning Representations, ICLR 2018 - Conference Track Proceedings}} (\bibinfo{year}{2018}), \bibinfo{pages}{1--13}.
\newblock
\showeprint[arxiv]{1710.09412}


\bibitem[Zhang et~al\mbox{.}(2020)]%
        {zhang2020does}
\bibfield{author}{\bibinfo{person}{Linjun Zhang}, \bibinfo{person}{Zhun Deng}, \bibinfo{person}{Kenji Kawaguchi}, \bibinfo{person}{Amirata Ghorbani}, {and} \bibinfo{person}{James Zou}.} \bibinfo{year}{2020}\natexlab{}.
\newblock \showarticletitle{How does mixup help with robustness and generalization?}
\newblock \bibinfo{journal}{\emph{arXiv preprint arXiv:2010.04819}} (\bibinfo{year}{2020}).
\newblock


\bibitem[Zhou et~al\mbox{.}(2020)]%
        {zhou2020hierarchy}
\bibfield{author}{\bibinfo{person}{Jie Zhou}, \bibinfo{person}{Chunping Ma}, \bibinfo{person}{Dingkun Long}, \bibinfo{person}{Guangwei Xu}, \bibinfo{person}{Ning Ding}, \bibinfo{person}{Haoyu Zhang}, \bibinfo{person}{Pengjun Xie}, {and} \bibinfo{person}{Gongshen Liu}.} \bibinfo{year}{2020}\natexlab{}.
\newblock \showarticletitle{Hierarchy-aware global model for hierarchical text classification}. In \bibinfo{booktitle}{\emph{Proceedings of the 58th Annual Meeting of the Association for Computational Linguistics}}. \bibinfo{pages}{1106--1117}.
\newblock


\bibitem[Zhou et~al\mbox{.}(2016)]%
        {textrnnattn}
\bibfield{author}{\bibinfo{person}{Peng Zhou}, \bibinfo{person}{Wei Shi}, \bibinfo{person}{Jun Tian}, \bibinfo{person}{Zhenyu Qi}, \bibinfo{person}{Bingchen Li}, \bibinfo{person}{Hongwei Hao}, {and} \bibinfo{person}{Bo Xu}.} \bibinfo{year}{2016}\natexlab{}.
\newblock \showarticletitle{{Attention-based bidirectional long short-term memory networks for relation classification}}.
\newblock \bibinfo{journal}{\emph{54th Annual Meeting of the Association for Computational Linguistics, ACL 2016 - Short Papers}} (\bibinfo{year}{2016}), \bibinfo{pages}{207--212}.
\newblock
\showISBNx{9781510827592}
\urldef\tempurl%
\url{https://doi.org/10.18653/v1/p16-2034}
\showDOI{\tempurl}


\bibitem[Zhu et~al\mbox{.}(2019)]%
        {zhu2019mixup}
\bibfield{author}{\bibinfo{person}{Yingke Zhu}, \bibinfo{person}{Tom Ko}, {and} \bibinfo{person}{Brian Mak}.} \bibinfo{year}{2019}\natexlab{}.
\newblock \showarticletitle{Mixup Learning Strategies for Text-Independent Speaker Verification.}. In \bibinfo{booktitle}{\emph{Interspeech}}. \bibinfo{pages}{4345--4349}.
\newblock


\bibitem[Zhu et~al\mbox{.}(2020)]%
        {10.1145/3397271.3401177}
\bibfield{author}{\bibinfo{person}{Ziwei Zhu}, \bibinfo{person}{Jianling Wang}, {and} \bibinfo{person}{James Caverlee}.} \bibinfo{year}{2020}\natexlab{}.
\newblock \showarticletitle{Measuring and Mitigating Item Under-Recommendation Bias in Personalized Ranking Systems}. In \bibinfo{booktitle}{\emph{Proceedings of the 43rd International ACM SIGIR Conference on Research and Development in Information Retrieval}} (Virtual Event, China) \emph{(\bibinfo{series}{SIGIR '20})}. \bibinfo{publisher}{Association for Computing Machinery}, \bibinfo{address}{New York, NY, USA}, \bibinfo{pages}{449–458}.
\newblock
\showISBNx{9781450380164}
\urldef\tempurl%
\url{https://doi.org/10.1145/3397271.3401177}
\showDOI{\tempurl}


\end{thebibliography}
